\documentclass[10pt,twocolumn,letterpaper]{article}

\usepackage{cvpr}              % To produce the CAMERA-READY version
\usepackage{times}

\usepackage[utf8]{inputenc} % allow utf-8 input
\usepackage[T1]{fontenc}    % use 8-bit T1 fonts
\usepackage{url}            % simple URL typesetting
\usepackage{booktabs}       % professional-quality tables
\usepackage{amsfonts}       % blackboard math symbols
\usepackage{nicefrac}       % compact symbols for 1/2, etc.
\usepackage{microtype}      % microtypography
\usepackage{xcolor}         % colors

\usepackage{graphicx}
\usepackage{amsmath,amssymb}
\usepackage{verbatim}
\usepackage{multirow}
\usepackage{marvosym}
\usepackage{makecell}
\usepackage{bm}
\usepackage{wrapfig}
\usepackage{caption}

\usepackage[accsupp]{axessibility}  % Improves PDF readability for those with disabilities.

\newtheorem{theorem}{Theorem}
\newtheorem{definition}{Definition}
\newtheorem{corollary}{Corollary}

\usepackage[pagebackref,breaklinks,colorlinks]{hyperref}

\usepackage[capitalize]{cleveref}
\crefname{section}{Sec.}{Secs.}
\Crefname{section}{Section}{Sections}
\Crefname{table}{Table}{Tables}
\crefname{table}{Tab.}{Tabs.}

\def\cvprPaperID{1763} % *** Enter the CVPR Paper ID here
\def\confName{CVPR}
\def\confYear{2023}

\begin{document}

\title{EqMotion: Equivariant Multi-agent Motion Prediction \\ with Invariant Interaction Reasoning}
\author{Chenxin Xu\textsuperscript{1,2}, Robby T. Tan\textsuperscript{2}, Yuhong Tan\textsuperscript{1}, Siheng Chen\textsuperscript{1,3\footnotemark[1]},  \\ Yu Guang Wang\textsuperscript{1}, Xinchao Wang\textsuperscript{2}, Yanfeng Wang\textsuperscript{3,1}
\\\textsuperscript{1}Shanghai Jiao Tong University,  
\textsuperscript{2}National University of Singapore,
\textsuperscript{3}Shanghai AI Laboratory
\\
{\tt\small {\{xcxwakaka,tyheeeer,sihengc,yuguang.wang,wangyanfeng\}@sjtu.edu.cn},} 
\tt\small {\{robby.tan,xinchao\}@nus.edu.sg}
}

\maketitle

\renewcommand{\thefootnote}{\fnsymbol{footnote}}
\footnotetext[1]{Corresponding author.}
% \footnotetext{Code is available at: \url{https://github.com/MediaBrain-SJTU}}

%%%%%%%%% ABSTRACT
\begin{abstract}

Learning to predict agent motions with relationship reasoning is important for many applications.
In motion prediction tasks, maintaining motion equivariance under Euclidean geometric transformations and invariance of agent interaction is a critical and fundamental principle. However, such equivariance and invariance properties are overlooked by most existing methods.
To fill this gap, we propose EqMotion, an efficient equivariant motion prediction model with invariant interaction reasoning. To achieve motion equivariance, we propose an equivariant geometric feature learning module to learn a Euclidean transformable feature through dedicated designs of equivariant operations. To reason agent's interactions, we propose an invariant interaction reasoning module to achieve a more stable interaction modeling. To further promote more comprehensive motion features, we propose an invariant pattern feature learning module to learn an invariant pattern feature, which cooperates with the equivariant geometric feature to enhance network expressiveness. 
We conduct experiments for the proposed model on four distinct scenarios: particle dynamics, molecule dynamics, human skeleton motion prediction and pedestrian trajectory prediction. Experimental results show that our method is not only generally applicable, but also achieves state-of-the-art prediction performances on all the four tasks, improving by $24.0/30.1/8.6/9.2\%$. 
Code is available at \url{https://github.com/MediaBrain-SJTU/EqMotion}.
\end{abstract}

% \tableofcontents

%%%%%%%%% BODY TEXT
\section{Introduction}

\begin{figure}[t] 
\centering
\includegraphics[width=0.47\textwidth]{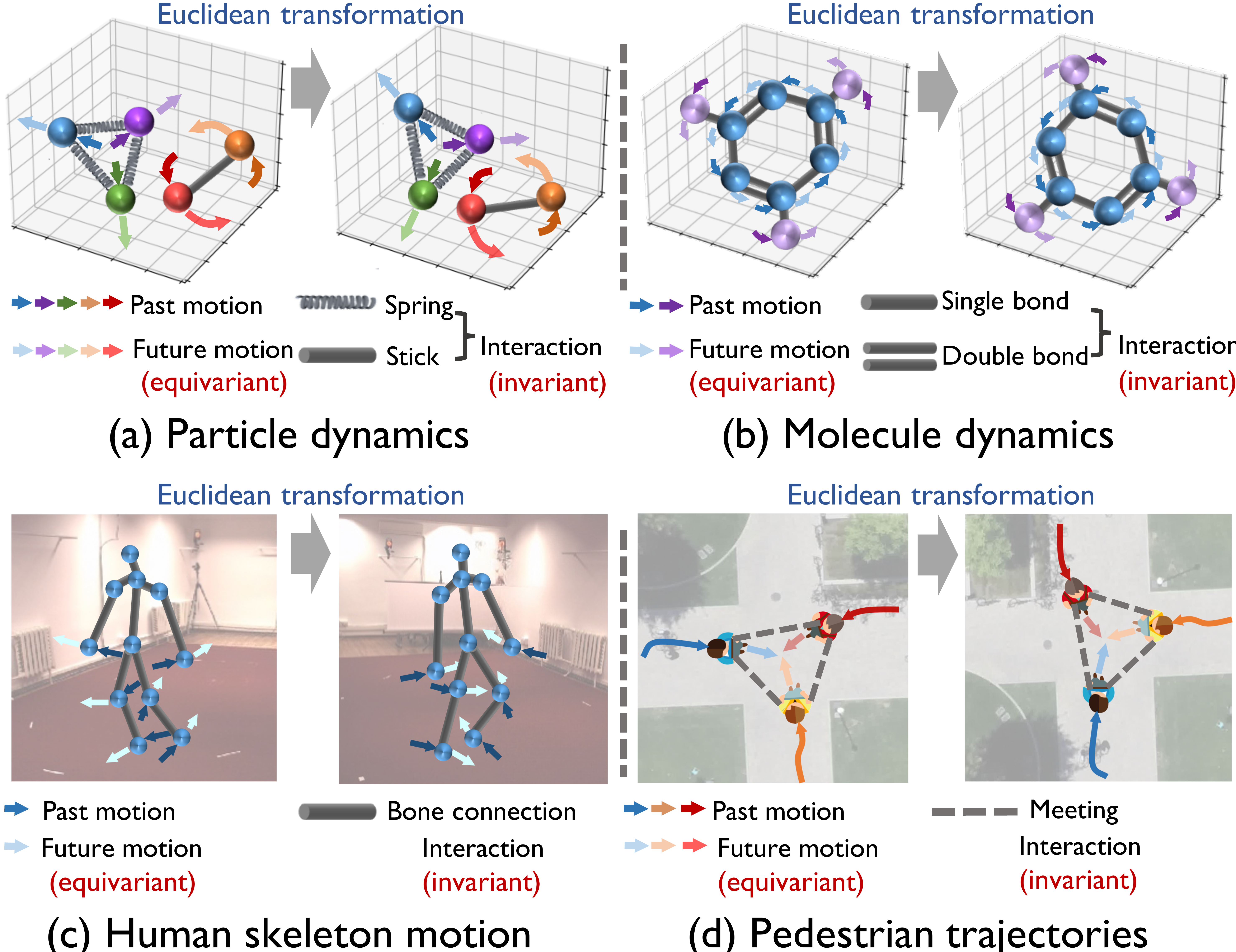}
\vspace{-3mm}
\caption{\small Motion equivariance and interaction invariance under the Euclidean
geometric transformation is a fundamental principle for a prediction model, but this principle is often overlooked by previous works. In this work, we propose EqMotion to fill this gap.}
\label{fig:jewel}
\vspace{-4mm}
\end{figure}

Motion prediction aims to predict future trajectories of multiple interacting agents given their historical observations. It is widely studied in many applications like physics \cite{battaglia2016interaction, kipf2018neural}, molecule dynamics \cite{chmiela2017machine}, autonomous driving \cite{levinson2011towards} and human-robot interaction \cite{li2019actional,xu2021invariant}.
In the task of motion prediction, an often-overlooked yet fundamental principle is that a prediction model is required to be equivariant under the Euclidean geometric transformation (including translation, rotation and reflection), and at the same time maintain the interaction relationships invariant. 
Motion equivariance here means that if an input motion is transformed under a Euclidean transformation, the output motion must be equally transformed under the same transformation. Interaction invariance means that the way agents interact remains unchanged under the input's transformation. Figure \ref{fig:jewel} shows real-world examples of motion equivariance and interaction invariance.

Employing this principle in a network design brings at least two benefits. First, the network will be robust to arbitrary Euclidean transformations. Second, the network will have the capability of being generalizable over rotations and translations of the data. This capability makes the network more compact, reducing the network's learning burden and contributing to a more accurate prediction. 

Despite the motion equivariance property being important and fundamental, it is often neglected and not guaranteed by most existing motion prediction methods. The main reason is that these methods transform the input motion sequence directly into abstract feature vectors, where the geometric transformations are not traceable, causing the geometric relationships between agents to be irretrievable. Random augmentation will ease the equivariance problem, but it is still unable to guarantee the equivariance property.
\cite{kofinas2021roto} uses non-parametric pre and post coordinate processing to achieve equivariance, but its parametric network structures do not satisfy equivariance. Some methods propose equivariant parametric network structures utilizing the higher-order representations of spherical harmonics \cite{thomas2018tensor,fuchs2020se} or proposing an equivariant message passing \cite{satorras2021n}, but they focus on the state-to-state prediction. This means that they use only one historical timestamp to predict one future timestamp. Consequently, these methods have limitations on utilizing motion's temporal information and modeling interaction relationships since a single-state observation is insufficient for both interaction modeling and temporal dependency modeling.

In this paper, we propose EqMotion, the first motion prediction model that is theoretically equivariant to the input motion under Euclidean geometric transformations based on the parametric network. The proposed EqMotion has three novel designs: equivariant geometric feature learning, invariant pattern feature learning and invariant interaction reasoning. 
To ensure motion equivariance, we propose an equivariant geometric feature learning module to learn a Euclidean transformable geometric feature through dedicated designs of equivariant operations. The geometric feature preserves motion attributes that are relevant to Euclidean transformations. To promote more comprehensive representation power, we introduce an invariant pattern feature learning module to complement the network with motion attributes that are independent of Euclidean transformations. The pattern features, cooperated into the geometric features, provide expressive motion representations by exploiting motions' spatial-temporal dependencies. 

To further infer the interactions during motion prediction, we propose an invariant interaction reasoning module, which ensures that the captured interaction relationships are invariant to the input motion under Euclidean transformations. 
The module infers an invariant interaction graph by utilizing invariant factors in motions. The edge weights in the interaction graph categorize agents' interactions into different types, leading to better interaction representation.

We conduct extensive experiments on four different scenarios to evaluate our method's effectiveness: particle dynamics, molecule dynamics, 3D human skeleton motion and pedestrian trajectories. Comparing to many task-specific motion prediction methods, our method is generally applicable and achieves state-of-the-art performance in all these tasks by reducing the prediction error by 24.0/30.1/8.6/9.2$\%$ respectively.
We also present that EqMotion is lightweight, and has a model size less than 30$\%$ of many other models' sizes. We show that EqMotion using only 5 $\%$ data can achieve a comparable performance with other methods that take full data. As a summary, here are our contributions:

% The main contributions of this paper are:

$\bullet$ We propose EqMotion, the first motion prediction model that theoretically ensures sequence-to-sequence motion equivariance based on the parametric network. With equivariance, EqMotion promotes more generalization ability of motion feature learning, leading to more robust and accurate prediction.

$\bullet$ We propose a novel invariant interaction reasoning module, in which the captured interactions between agents are invariant to the input motion under Euclidean geometric transformations. With this, EqMotion achieves more generalization ability and stability in the interaction reasoning.

$\bullet$ We conduct experiments on four types of scenarios and find that EqMotion is applicable to all these different 
tasks, and importantly outperforms existing state-of-the-art methods on all the tasks.

\label{sec:intro}

\section{Related Work}
\vspace{-1mm}
\textbf{Equivariant Networks.} Equivariance first draws high attention on the 2D image domain. Since CNN structure is sensitive to rotations, researchers start to explore rotation-equivariant designs like oriented convolutional filters \cite{cohen2016group,marcos2017rotation}, log-polar transform \cite{esteves2017polar}, circular harmonics \cite{worrall2017harmonic} or steerable filters \cite{weiler2018learning}. Meanwhile, GNN architectures~\cite{yang2020CVPR,yang2020NeurIPS,yang2022deep} exploring symmetries on both rotation and translation have been emerged. Specifically, \cite{ummenhofer2019lagrangian,sanchez2020learning} achieves partial symmetries by promoting translation equivariance. \cite{thomas2018tensor,fuchs2020se} builds filters using spherical harmonics allowing transformations between high-order representations, achieving the rotation and translation equivariance. 
% \cite{fuchs2020se} further extends the model with an attention mechanism containing invariant attention weights. 
\cite{finzi2020generalizing,hutchinson2021lietransformer} construct a Lie convolution to parameterize transformations into Lie algebra form. \cite{deng2021vector} proposes a series of equivariant layers for point cloud networks. \cite{jing2020learning} propose geometric vector perceptions for protein structure learning. Recently, EGNN \cite{satorras2021n} proposes a simple equivariant message passing form without using computationally expensive high-order representations. \cite{huang2022equivariant} further extends it by considering geometrical constraints. 
However, most existing methods are only applicable to state prediction, limiting 
models from exploiting sequence information. 
\cite{kofinas2021roto} uses pre and post coordinate processing to achieve motion equivariance but its network structure does not satisfy equivariance.
In this work, we propose an equivariant model based on the parametric network which is generally applicable to 
motion prediction tasks 
and achieves a more precise prediction.

\textbf{Motion Prediction.}
Motion prediction has wide application scenarios. \cite{battaglia2016interaction,mrowca2018flexible,sanchez2019hamiltonian} proposes graph neural networks for learning to simulate complex physical systems. \cite{kipf2018neural,graber2020dynamic,li2020evolvegraph,xu2022dynamic} both explicitly infer the interactions relationships and perform prediction in physical systems. For human or vehicle trajectory prediction, social forces \cite{helbing1995social,mehran2009abnormal}, Markov \cite{kitani2012activity,wang2007gaussian} process, and RNNs \cite{alahi2016social,morton2016analysis,vemula2018social} are first methods to employ. Multi-model prediction methods are further proposed like using generator-discriminator structures \cite{gupta2018social,hu2020collaborative}, multi-head output \cite{liang2020learning,tang2021collaborative}, conditional variational autoencoders \cite{mangalam2020not,lee2017desire,salzmann2020trajectron++,yuan2021agentformer,xu2022groupnet,xu2022dynamic,yang2022KF}, memory mechanisms \cite{xu2022remember,marchetti2020mantra}, Gaussian mixture distribution prediction \cite{graber2020dynamic,li2020evolvegraph}. The HD map information is specifically considered in the autonomous driving scenario \cite{hu2020collaborative,chai2019multipath,liang2020garden,casas2018intentnet,gao2020vectornet,liang2020learning,zhong2022aware}. For 3D human skeleton motion prediction, early methods are based on the state prediction \cite{lehrmann2014efficient,taylor2009factored}. Later RNN-based models considering the sequential motion states are proposed \cite{fragkiadaki2015recurrent,walker2017pose,jain2016structural,martinez2017human}. \cite{guo2019human,li2018convolutional} use spatial graph convolutions to directly regress the whole sequences. Besides, some methods \cite{mao2019learning, cai2020learning,mao2020history,li2022skeleton} specifically exploit the correlations between body joints. Multi-scale graphs are built by \cite{li2020dynamic,dang2021msr,li2021symbiotic} to capture different body-level dependencies. In this work, we propose a generally applicable motion prediction network which promotes Euclidean equivariance, a fundamental property but neglected by previous methods, to have a more robust and accurate prediction. 

\begin{figure*}[t] 
\centering
\includegraphics[width=0.96\textwidth]{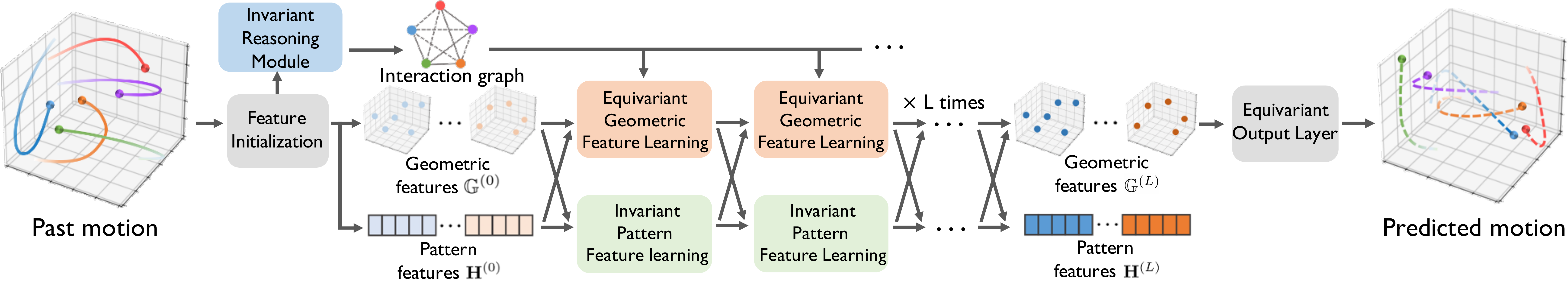}
\vspace{-3.5mm}
\caption{\small EqMotion architecture. In EqMotion, we first 
use a feature initialization layer to initialize geometric features and pattern features. We then successively update the geometric features and the pattern features by the equivariant geometric feature learning and invariant pattern feature learning layers, obtaining expressive feature representation. We further propose an invariant reasoning module to infer an interaction graph used in equivariant geometric feature learning. Finally, we use an equivariant output layer to obtain the final prediction.}
\label{fig:network}
\vspace{-3mm}
\end{figure*}

\vspace{-1mm}
\section{Background and Problem Formulation}
% In this section, we introduce the relevant definition of equivariance and the formulation of motion prediction.
\vspace{-1mm}
\subsection{Motion Prediction}
\vspace{-1mm}
Here we introduce the general problem formulation of motion prediction, which aims to generate future motions given the historical observations. Mathematically, consider $M$ agents in a multi-agent system. Let $\mathbf{X}_i =[\mathbf{x}_i^1,\mathbf{x}_i^2,\cdots \mathbf{x}_i^{T_{\rm p}}]\in \mathbb{R}^{T_{\rm p}\times n}$ and $\mathbf{Y}_i =[\mathbf{y}_i^1,\mathbf{y}_i^2,\cdots \mathbf{y}_i^{T_{\rm f}}] \in \mathbb{R}^{T_{\rm f}\times n}$ be the $i$th agent's past and future motion, where $T_{\rm p}$ and $T_{\rm f}$ are the past and future timestamps and $n$ is the dimension of the system space. $n$ usually equals to 2 or 3 (corresponding to 2D or 3D case). The $t$th timestamp locations $\mathbf{x}_i^t$ and $\mathbf{y}_i^t$ are $n$-dimension vectors. The whole system's past and future motion is represented as $\mathbb{X} =[\mathbf{X}_1,\mathbf{X}_2,\cdots,\mathbf{X}_M]\in \mathbb{R}^{M\times T_{\rm p}\times n}$ and $\mathbb{Y} = [\mathbf{Y}_1,\mathbf{Y}_2,\cdots,\mathbf{Y}_M] \in \mathbb{R}^{M\times T_{\rm f}\times n}$. We aim to propose a prediction network $\mathcal{F}_{\rm pred}(\cdot)$ so that the predicted future motions $\widehat{\mathbb{Y}} = \mathcal{F}_{\rm pred}(\mathbb{X})$ are as close to the ground-truth future motions $\mathbb{Y}$ as possible.

\subsection{Equivariance and Invariance}
\vspace{-1mm}
The Euclidean geometric transformation has three basic forms: translation, rotation and reflection. The translation is modeled by a translation vector and rotation (or reflection) is modeled by an orthogonal rotation (or reflection) matrix. Let $\mathbf{t} \in \mathbb{R}^n$ be a translation vector  and $\mathbf{R} \in {\rm SO}(n)$ be a $n\times n$ rotation  matrix, we have the following definitions for equivariance and invariance:
\vspace{-1.5mm}
\begin{definition}
\label{def:equivariance}
Let $\mathbb{X}$ be an input, $\mathcal{F}(\cdot)$ be an operation and $\mathbb{Z}=\mathcal{F}(\mathbb{X})$ be the corresponding output. The operation $\mathcal{F}(\cdot)$ is called equivariant under Euclidean transformation if
\begin{equation*}
   \setlength{\abovedisplayskip}{2pt}
   \setlength{\belowdisplayskip}{2pt}
\mathbb{Z} \mathbf{R}+\mathbf{t} = \mathcal{F}(\mathbb{X}\mathbf{R}+\mathbf{t})\quad \forall \,\mathbf{R} \in {\rm SO}(n), \forall \, \mathbf{t}\in \mathbb{R}^{n}.
\end{equation*}
\vspace{-3mm}
\end{definition}
\begin{definition}
\vspace{-4mm}
\label{def:invariance}
Let $\mathbb{X}$ be an input, $\mathcal{F}(\cdot)$ be an operation and $\mathbb{Z}=\mathcal{F}(\mathbb{X})$ be the corresponding output. The operation $\mathcal{F}(\cdot)$ is called invariant under Euclidean transformation if
\begin{equation*}
   \setlength{\abovedisplayskip}{2pt}
   \setlength{\belowdisplayskip}{0pt}
\mathbb{Z} = \mathcal{F}(\mathbb{X}\mathbf{R}+\mathbf{t})\quad \forall \,\mathbf{R} \in {\rm SO}(n), \forall \, \mathbf{t}\in \mathbb{R}^{n}.
\end{equation*}
\end{definition}
\vspace{-2mm}
Here, when a motion prediction network $\mathcal{F}_{\rm pred}(\cdot)$ satisfies Definition~\ref{def:equivariance}, $\mathcal{F}_{\rm pred}(\cdot)$ is said motion equivariant. When an interaction reasoning model $\mathcal{F}_{\rm reason}(\cdot)$ satisfies Definition~\ref{def:invariance}, $\mathcal{F}_{\rm reason}(\cdot)$ is considered interaction invariant. We also say the output $\mathbb{Z}$ is equivariant/invariant (to the input motion under Euclidean transformation) if Definition 
\ref{def:equivariance}/\ref{def:invariance} is satisfied.

In the following section, we will introduce our motion prediction network along with geometric features that are equivariant under Euclidean transformations, and pattern features along with the interaction reasoning module are invariant under Euclidean transformations.

\section{Methodology} 
\vspace{-1mm}
In this section, we present EqMotion, a motion prediction network which is equivariant under Euclidean geometric transformations. The whole network architecture is shown in Figure \ref{fig:network}. The core of EqMotion is to successively learn equivariant geometric features and invariant pattern features by mutual cooperation in designed equivariant/invariant operations, which not only provide expressive motion and interaction representations, but also preserve equivariant/invariant properties.
For agents' motions $\mathbb{X}\in \mathbb{R}^{M\times T_{\rm p}\times n}$, the overall procedure of the proposed EqMotion is formulated as
\begin{subequations}
   \setlength{\abovedisplayskip}{2pt}
   \setlength{\belowdisplayskip}{2pt}
    \begin{align}
       & \mathbb{G}^{(0)}, \mathbf{H}^{(0)} = \mathcal{F}_{\rm IL}(\mathbb{X}), \label{eq:all1}\\
       &  \{\mathbf{c}_{ij} \} = \mathcal{F}_{\rm IRM}(\mathbb{G}^{(0)}, \mathbf{H}^{(0)}), \label{eq:all4}\\
       & \mathbb{G}^{(\ell+1)} = \mathcal{F}_{\rm EGFL}^{(\ell)}(\mathbb{G}^{(\ell)}, \mathbf{H}^{(\ell)},\{\mathbf{c}_{ij} \}), \label{eq:all2}\\
       & \mathbf{H}^{(\ell+1)} = \mathcal{F}_{\rm IPFL}^{(\ell)}(\mathbb{G}^{(\ell)}, \mathbf{H}^{(\ell)}), \label{eq:all3}\\
       & \widehat{\mathbb{Y}} = \mathcal{F}_{\rm EOL}(\mathbb{G}^{(L)}). \label{eq:all5}
    \end{align}
\end{subequations}
Step (\ref{eq:all1}) uses an initialization layer $\mathcal{F}_{\rm IL}(\cdot)$ to obtain initial geometric features $\mathbb{G}^{(0)}$ and pattern features $\mathbf{H}^{(0)}$. 
For interaction relationship unavailable cases, Step (\ref{eq:all4}) uses an invariant interaction reasoning module $\mathcal{F}_{\rm IRM}(\cdot)$ to infer an interaction graph $\{\mathbf{c}_{ij} \}$ whose edge weight $\mathbf{c}_{ij}$ is the interaction category between agent $i$ and $j$.
Step (\ref{eq:all2}) uses the $\ell$th equivariant geometric feature learning layer $\mathcal{F}_{\rm EGFL}^{(\ell)}(\cdot)$ to learn the $(\ell$$+$$1)$th geometric feature $\mathbb{G}^{(\ell+1)}$. 
Step (\ref{eq:all3}) uses the $\ell$th invariant pattern feature learning layer $\mathcal{F}^{(\ell)}_{\rm IPFL}(\cdot)$ to learn the $(\ell$$+$$1)$th pattern feature $\mathbf{H}^{(\ell+1)}$. Step (\ref{eq:all2}) and Step (\ref{eq:all3}) will repeat $L$ times. Step (\ref{eq:all5}) uses an equivariant output layer $\mathcal{F}_{\rm EOL}(\cdot)$ to obtain the final prediction $\widehat{\mathbb{Y}}$.

Note that i) to incorporate the geometric feature's equivariance, we need to design equivariant operations for the initialization layer $\mathcal{F}_{\rm IL}(\cdot)$ and the equivariant 
geometric feature learning layer $\mathcal{F}_{\rm EGFL}^{(\ell)}(\cdot)$; ii) to introduce the pattern feature's invariance, we need to design invariant operations for the initialization layer $\mathcal{F}_{\rm IL}(\cdot)$ and the invariant  
pattern feature learning layer $\mathcal{F}_{\rm IPFL}^{(\ell)}(\cdot)$; and iii) the interaction graph categorizes agent's spatial interaction into different categories for better interaction representing. The interaction graph is invariant  due to the reasoning module $\mathcal{F}_{\rm IRM}(\cdot)$ design. 
In subsequent sections, we elaborate the details of each step.

\subsection{Feature Initialization}
\label{sec:init}
\vspace{-1mm}
The feature initialization layer aims to obtain initial geometric features and pattern features while equipping them with different functionality. 
The initial geometric feature is denoted as $\mathbb{G}^{(0)} =[\mathbf{G}_1^{(0)},\cdots,\mathbf{G}_M^{(0)}] \in \mathbb{R}^{M\times C\times n}$ whose $i$th agent's geometric feature consists of $C$ geometric coordinates. The initial pattern feature is denoted as 
$\mathbf{H}^{(0)}=[\mathbf{h}_1^{(0)},\cdots,\mathbf{h}_M^{(0)}] \in \mathbb{R}^{M\times D}$ whose $i$th agent's pattern feature is a $D$-dimensional vector. Given the past motions $\mathbb{X}$ whose the $i$th agent's motion is $\mathbf{X}_i\in \mathbb{R}^{T_{\rm p}\times n}$, we obtain two initial features of the $i$th agent: 
\begin{equation*}
    \setlength{\abovedisplayskip}{2pt}
   \setlength{\belowdisplayskip}{2pt}
    \begin{aligned}
        &\mathbf{G}_i^{(0)} = \phi_{\rm init\_g}(\mathbf{X}_i-\overline{\mathbb{X}})+\overline{\mathbb{X}},\\
          &\mathbf{V}_i=\bigtriangleup \mathbf{X}_i ,\rho_i^t = ||\mathbf{V}_i^t||_2, \theta_{i}^t = \mathrm{angle}(\mathbf{V}_i^t,\mathbf{V}_i^{t-1}), \\
   & \mathbf{h}_i^{(0)}=\phi_{\mathrm {init\_h}}([\, \rho_i;\theta_i\,]),
    \end{aligned}
\end{equation*}
where $\phi_{\rm init\_g}(\mathbf{X})=\mathbf{W}_{\rm init\_g}\mathbf{X}, \mathbf{W}_{\rm init\_g} \in \mathbb{R}^{C \times {T_p}}$ is a linear function, $\overline{\mathbb{X}}$ is the mean coordinate of all agents all past timestamps. $\bigtriangleup$ is the difference operator to obtain the velocity $\mathbf{V}_i \in \mathbb{R}^{T_{\rm p} \times n}$.
$\mathbf{\rho}_i \in \mathbb{R}^{T_{\rm p}}$ is the velocity magnitude sequence and $\mathbf{\theta}_{i}\in \mathbb{R}^{T_{\rm p}}$ is the velocity angle sequence. The superscript $(\cdot)^{t}$ denotes the $t$th element, $||\cdot||_2$ is vector 2-norm, $\mathrm{angle}(\cdot)$ is the function calculating the angle between two vectors and $\phi_{\mathrm {init\_h}}(\cdot)$ is an embedding function implemented by MLP or LSTM. 
$[\cdot\,;\,\cdot]$ represents concatenation.

The geometric feature preserves both equivariant property and motion attributes that are sensitive to Euclidean geometric transforms since we linearly combine the equivariant locations. 
The pattern feature remains invariant and is sensitive to motion attributes independent of Euclidean geometric transforms.

\subsection{Invariant Reasoning Module}
\vspace{-1mm}
\label{sec:reason}
For most scenarios, the interaction relationship is implicit and unavailable. 
Thus, we propose an invariant reasoning module for inferring the interaction category between agents. Note that we design the reasoning module to be invariant as the interaction category is independent of Euclidean transformations. The output of the reasoning module is an invariant interaction graph whose edge weight $\mathbf{c}_{ij}\in [0,1]^K$ is a categorical vector representing the type of interaction between agent $i$ and $j$. $K$ is the interaction category number. To obtain the interaction categorical vector, we perform a message passing operation using the agent's initial pattern feature $\mathbf{h}_i^{(0)}$ and geometric feature $\mathbf{G}_i^{(0)}$:
\begin{equation*}
       \setlength{\abovedisplayskip}{2pt}
   \setlength{\belowdisplayskip}{2pt}
\begin{aligned}
&\mathbf{m}^{\prime}_{ij}=\phi_{\rm rm}\big([\mathbf{h}_i^{(0)};\mathbf{h}_j^{(0)};||\mathbf{G}_i^{(0)}-\mathbf{G}_j^{(0)}||_{2,\mathrm {col}}]\big),\\
&\mathbf{p}^{\prime}_i = \sum_{j \in \mathcal{N}_i} \mathbf{m}^{\prime}_{ij},\;\; \mathbf{h}_i^{\prime}=\phi_{\rm rh}\big([\mathbf{p}_i^{\prime};\mathbf{h}_i^{(0)}]\big),\\
&\mathbf{c}_{ij} = \mathrm{sm}\Big(\phi_{\rm rc}\big([\mathbf{h}_i^{\prime};\mathbf{h}_j^{\prime};||\mathbf{G}_i^{(0)}-\mathbf{G}_j^{(0)}||_{2,\mathrm {col}}]\big)/\tau\Big),
\end{aligned}
\end{equation*}
where $||\cdot ||_{2,\mathrm{col}}$ is a column-wise $\ell_2$-distance, $\phi_{\rm rm}(\cdot)$, $\phi_{\rm rh}(\cdot)$, $\phi_{\rm rc}(\cdot)$ are learnable functions implemented by MLPs. $\mathrm{sm}(\cdot)$ is the softmax function and $\tau$ is the temperature to control the smoothness of the categorical distribution. The interaction edge weights $\{\mathbf{c}_{ij}\}$ will be end-to-end learned with the whole prediction network. 
% The reasoned interaction category is invariant because of the invariance of the initial pattern features and relative $\ell_2$ distance between initial geometric features.

\subsection{Equivariant Geometric Feature Learning}
\label{sec:geo update}
\vspace{-1mm}
The equivariant geometric feature learning process aims to find more representative agents' geometric features by exploiting their spatial and temporal dependencies, while maintaining the equivariance. 
% During the process, the pattern feature plays a complementing role by providing geometric features with invariant movement pattern information that is unrelated to Euclidean transformations. 
During the process, we perform i) equivariant inner-agent attention to exploit the temporal dependencies; ii) equivariant inter-agent aggregation to model spatial interactions; and iii) equivariant non-linear function to further enhance the representation ability.

\textbf{Equivariant inner-agent attention}
To exploit the temporal dependency, we perform an attention mechanism on the coordinate dimension of the geometric feature since the coordinate dimension originates from the temporal dimension. 
Given the $i$th agent's geometric feature $\mathbf{G}^{(\ell)}_i \in \mathbb{R}^{C\times n}$ and pattern feature $\mathbf{h}_i^{(\ell)}$, we have  
\begin{equation}
        \setlength{\abovedisplayskip}{2pt}
   \setlength{\belowdisplayskip}{2pt}
   \begin{aligned}
       \mathbf{G}_i^{(\ell)} \leftarrow \phi^{(\ell)}_{\rm att}(\mathbf{h}_i^{(\ell)}) \cdot (\mathbf{G}_i^{(\ell)}-\overline{\mathbb{G}}^{(\ell)})+\overline{\mathbb{G}}^{(\ell)},
   \end{aligned}
\label{eq:inner}
\end{equation}
where $\phi^{(\ell)}_{\rm att}(\cdot): \mathbb{R}^{D}\rightarrow \mathbb{R}^{C}$ is a function which learns the attention weight for per coordinate, and $\overline{\mathbb{G}}^{(\ell)}$ is the mean coordinates summing up over all agents and all coordinates.

The above operation will bring the following benefits: i) The learned attention weight is invariant and the learned dependencies between different timestamps of the motion will not be disturbed by irrelevant Euclidean transformations; and ii) The learned geometric feature is equivariant because of the coordinate-wise linear multiplication.  

\textbf{Equivariant inter-agent aggregation}
% \label{subsec:inter geo}
The equivariant inter-agent aggregation aims to model spatial interactions between agents. The key idea is to use the reasoned or provided interaction category to learn aggregation weights, and then use the weights to aggregate neighboring agents' geometric features. The aggregation operation reads
\begin{equation}
\setlength{\abovedisplayskip}{0pt}
   \setlength{\belowdisplayskip}{0pt}
\begin{small}
   \begin{aligned}
&\mathbf{e}_{ij}^{(\ell)} = \sum_{k=1}^{K}\mathbf{c}_{ij,k}\phi^{(\ell)}_{e,k}([\mathbf{h}_i^{(\ell)};\mathbf{h}_j^{(\ell)};||\mathbf{G}^{(\ell)}_i-\mathbf{G}^{(\ell)}_j||_{2,\mathrm{col}}]),  \\
&\mathbf{G}^{(\ell)}_i \leftarrow \mathbf{G}^{(\ell)}_i + \sum_{j \in \mathcal{N}_i} \mathbf{e}_{ij}^{(\ell)} \cdot (\mathbf{G}^{(\ell)}_i-\mathbf{G}^{(\ell)}_j), 
\end{aligned}
\end{small}
\label{eq:inter}
\end{equation}
where $\mathbf{e}_{ij}^{(\ell)} \in \mathbb{R}^{C}$ is the learned aggregation weights between agent $i$ and $j$, $||\cdot ||_{2,\mathrm{col}}$ is a column-wise $\ell_2$-distance, $\cdot$ is dot product and $\mathcal{N}_i$ is the $i$th agent's neighboring set. For the $k$th interaction category, we assign a function $\phi^{(\ell)}_{e,k}(\cdot)$ that is implemented by MLP to model how the interaction works. 

The social influence from agent $j$ to agent $i$ is modeled by the coordinates' difference between the two agents. The intuition behind this design is that the mutual force between two objects is always in the direction of the line they formed. 
% The aggregation operation is equivariant since the coordinates' relative difference is equivariant with rotations and invariant to translations, and the input geometric feature is equivariant, leading to their sum being equivariant.

Note that for most scenarios, we do not have an explicit definition of interacted "neighbors", thus we use a fully-connected graph structure, which means every agent's neighbors include all other agents. For some special cases with massive nodes, like point clouds, we construct a local neighboring set by choosing neighbors within a distance threshold.

\textbf{Equivariant non-linear function}
According to Eq.~(\ref{eq:inner}) and Eq.~\eqref{eq:inter}, the operation of inner-agent attention and inter-agent aggregation are linear combinations with agents' coordinates. It is well known that non-linear operation is the key to improving neural networks' expressivity. Therefore, we propose an equivariant non-linear function to enhance the representation ability of our prediction network while preserving equivariance. The key idea is proposing a criterion with the invariance property for splitting different conditions, 
and for each condition design an equivariant equation.
Mathematically, the non-linear function is
\begin{equation*}
        \setlength{\abovedisplayskip}{2pt}
   \setlength{\belowdisplayskip}{2pt}
\begin{footnotesize}
   \begin{aligned}
&\mathbf{Q}_i^{(\ell)} = \mathbf{W}_{\rm Q}^{(\ell)}(\mathbf{G}_i^{(\ell)}-\overline{\mathbb{G}}^{(\ell)}), \; 
    \mathbf{K}_i^{(\ell)} = \mathbf{W}_{\rm K}^{(\ell)}(\mathbf{G}_i^{(\ell)}-\overline{\mathbb{G}}^{(\ell)}), \\
        &\mathbf{g}_{i,c}^{(\ell+1)}=
        \begin{cases}
        \mathbf{q}_{i,c}^{(\ell)}     +\overline{\mathbb{G}}^{(\ell)}        & 	\text{if} \; \langle \mathbf{q}_{i,c}^{(\ell)}, \mathbf{k}_{i,c}^{(\ell)}\rangle \geq 0,\\
        \mathbf{q}_{i,c}^{(\ell)} - \langle \mathbf{q}_{i,c}^{(\ell)}, \frac{\mathbf{k}_{i,c}^{(\ell)}}{||\mathbf{k}_{i,c}^{(\ell)}||_2}\rangle \frac{\mathbf{k}_{i,c}^{(\ell)}}{||\mathbf{k}_{i,c}^{(\ell)}||_2} +\overline{\mathbb{G}}^{(\ell)}       & \text{otherwise},
        \end{cases}
\end{aligned}
\end{footnotesize}
\end{equation*}
where $\mathbf{Q}_{i}^{(\ell)}$ and  $\mathbf{K}_{i}^{(\ell)}$ is the learned query coordinates and key coordinates, $\mathbf{W}_{\rm K}^{(\ell)} \in \mathbb{R}^{C\times C}$ are learnable matrices for queries and keys. $\overline{\mathbb{G}}^{(\ell)}$ is the mean coordinates over all agents and all coordinates. $\mathbf{q}_{i,c}^{(\ell)}$, $\mathbf{k}_{i,c}^{(\ell)}$ is the $c$th coordinate of $\mathbf{Q}_{i}^{(\ell)}$ and $\mathbf{K}_{i}^{(\ell)}$. $\langle \cdot, \cdot \rangle$
is the vector inner product.

For every geometric coordinate $\mathbf{g}_{i,c}^{(\ell)}$, we learn a query coordinate $\mathbf{q}_{i,c}^{(\ell)}$ and a key coordinate $\mathbf{k}_{i,c}^{(\ell)}$. We set the criterion as the inner product of the query coordinate and the key coordinate. If the inner product is positive, we directly take the value of the query coordinate as output; otherwise, we clip the query coordinate vector by moving out its parallel components with the key coordinate vector. Finally, we obtain the 
$i$th agent's geometric features of the next layer $\mathbf{G}_i^{(l+1)}$ by gathering all the coordinates $\mathbf{g}_{i,c}^{(l+1)}$. The non-linear function is equivariant, since the criterion is invariant and the two equations under two conditions are equivariant. 

\subsection{Invariant Pattern Feature Learning}
\label{sec:pattern update}
\vspace{-1mm}
The invariant pattern feature learning aims to obtain a more representative agent pattern feature through interacting with neighbors. Here we learn the agent's pattern feature with an invariant message passing mechanism. Specially, we add the relative geometric feature difference into the edge modeling in the message passing to complement the information of relationships between agent absolute locations, which cannot be obtained by solely using the pattern features. Given the $i$th agent's pattern feature $\mathbf{h}_i^{(\ell)}\in \mathbb{R}^{D}$ and geometric feature $\mathbf{G}_i^{(\ell)} \in \mathbb{R}^{C\times n}$, the next layer's pattern feature $\mathbf{h}_i^{(l+1)}$ is obtained by
\begin{equation*}
        \setlength{\abovedisplayskip}{2pt}
   \setlength{\belowdisplayskip}{0pt}
\begin{aligned}
    &\mathbf{m}_{ij}^{(\ell)} = \phi_m^{(\ell)}([\mathbf{h}_i^{(\ell)};\mathbf{h}_j^{(\ell)};||\mathbf{G}_i^{(\ell)}-\mathbf{G}_j^{(\ell)}||_{2,\mathrm {col}}]), \\[1mm]
    &\mathbf{p}_{i}^{(\ell)} = \sum_{j \in \mathcal{N}(i)}\mathbf{m}_{ij}^{(\ell)},
    \;\mathbf{h}_i^{(l+1)} = \phi_h^{(\ell)}([\mathbf{h}_i^{(\ell)};\mathbf{p}_{i}^{(\ell)}]).
\end{aligned}
\end{equation*}
Here $\mathbf{m}_{ij}^{(\ell)}$ is the edge features and $\mathbf{p}_{i}^{(\ell)}$ is the aggregated neighboring features in the message passing. $||\cdot ||_{2,\mathrm{col}}$ is a column-wise $\ell_2$-distance and $[\cdot \,; \, \cdot]$ denotes concatenation. Functions $\phi_m^{(\ell)}(\cdot)$ and $\phi_h^{(\ell)}(\cdot)$ are implemented by MLPs.

\subsection{Equivariant Output Layer}
\label{sec:output}
\vspace{-1mm}
After total $L$ feature learning layers, we obtain the final $i$th agent's geometric feature $\mathbf{G}_i^{(L)}$ and pattern feature $\mathbf{h}_i^{(L)}$. We use the geometric feature for the final prediction by a linear operation for equivariance: for the $i$th agent:
\begin{equation*}
 \setlength{\abovedisplayskip}{2pt}
   \setlength{\belowdisplayskip}{2pt}
    \widehat{\mathbf{Y}_i} = \mathbf{W}_{\rm out}(\mathbf{G}_i^{(\ell)}-\overline{\mathbb{G}}^{(\ell)})+\overline{\mathbb{G}}^{(\ell)},
\end{equation*}
where $\mathbf{W}_{\rm out} \in \mathbb{R}^{T_{\rm f}\times D}$ is a learnable weight matrix. Finally, we gather all the agent prediction $\widehat{\mathbf{Y}_i}$ to have the 
final predicted motions $\widehat{\mathbb{Y}}$.

\subsection{Theoretical Analysis}
\label{sec:theory}
\vspace{-1mm}
In this section, we analyze our network's equivariance property and the interaction reasoning module's invariance property, as stated in the following theorem. Let $\mathbb{G}^{(\ell)}\in \mathbb{R}^{M\times C \times n}$ be all agents' geometric features, $\mathbf{H}^{(\ell)}\in \mathbb{R}^{M\times D}$ be all agents' pattern features at the $\ell$th layer, and $\{\mathbf{c}_{ij} \}$ be the set of all the interaction categorical vectors.  
\begin{theorem}
\vspace{-1mm}
\label{theorem:1}
For arbitrary translation vector $\mathbf{t} \in \mathbb{R}^{n}$ and rotation (or reflection) matrix $\mathbf{R} \in {\rm SO}(n)$, the modules with equivariance and invariance in our network satisfy:

1. For the initialization layer $\mathcal{F}_{\rm IL}(\cdot)$, the initial geometric feature is equivariant and the initial pattern feature is invariant:
\begin{equation*}
       \setlength{\abovedisplayskip}{2pt}
   \setlength{\belowdisplayskip}{2pt}
    \mathbb{G}^{(0)}\mathbf{R}+ \mathbf{t},\, \mathbf{H}^{(0)} = \mathcal{F}_{\rm IL}(\mathbb{X}\mathbf{R} + \mathbf{t}).
\end{equation*}

2. The reasoning module $\mathcal{F}_{\rm IRM}(\cdot)$ along with reasoned interaction categorical vectors $\{\mathbf{c}_{ij} \}$ is invariant:
\begin{equation*}
       \setlength{\abovedisplayskip}{2pt}
   \setlength{\belowdisplayskip}{2pt}
    \{\mathbf{c}_{ij} \} = \mathcal{F}_{\rm IRM}(\mathbb{G}^{(0)}\mathbf{R}+ \mathbf{t}, \mathbf{H}^{(0)}).
\end{equation*}

3. The $\ell$th geometric feature learning layer $\mathcal{F}_{\rm EGFL}^{(\ell)}(\cdot)$ is equivariant:
\begin{equation*}
       \setlength{\abovedisplayskip}{2pt}
   \setlength{\belowdisplayskip}{2pt}
\mathbb{G}^{(l+1)}\mathbf{R}+ \mathbf{t} = \mathcal{F}_{\rm EGFL}^{(\ell)}(\mathbb{G}^{(\ell)}\mathbf{R}+ \mathbf{t}, \mathbf{H}^{(\ell)},\{\mathbf{c}_{ij} \}).
\end{equation*}

4. The $\ell$th pattern feature learning layer $\mathcal{F}_{\rm IPFL}^{(\ell)}(\cdot)$ is invariant:
\begin{equation*}
       \setlength{\abovedisplayskip}{2pt}
   \setlength{\belowdisplayskip}{2pt}
\mathbf{H}^{(l+1)} = \mathcal{F}_{\rm IPFL}^{(\ell)}(\mathbb{G}^{(\ell)}\mathbf{R}+ \mathbf{t}, \mathbf{H}^{(\ell)}).
\end{equation*}

5. The output layer $\mathcal{F}_{\rm EOL}(\cdot)$ is equivariant:
\begin{equation*}
       \setlength{\abovedisplayskip}{2pt}
   \setlength{\belowdisplayskip}{0pt}
\widehat{\mathbb{Y}}\mathbf{R}+ \mathbf{t} = \mathcal{F}_{\rm EOL}(\mathbb{G}^{(L)}\mathbf{R}+ \mathbf{t}).
\end{equation*}
\vspace{-7mm}
\end{theorem}
See the detailed proof in Appendix. Theorem \ref{theorem:1} presents the equivariance/invariance properties with the Euclidean transformation for each operation in the proposed network. Based on Theorem~\ref{theorem:1}, by combining all network operations together, we can show that the whole network is equivariant:
\begin{corollary}
\vspace{-2mm}
\label{cor:equivariance}
For arbitrary translation vector $\mathbf{t} \in \mathbb{R}^{n}$ and rotation matrix $\mathbf{R} \in {\rm SO}(n)$, our whole network EqMotion $\mathcal{F}_{\rm pred}(\cdot)$ satisfies:
\begin{equation*}
        \setlength{\abovedisplayskip}{2pt}
   \setlength{\belowdisplayskip}{2pt}
\widehat{\mathbb{Y}}\mathbf{R}+\mathbf{t} = \mathcal{F}_{\rm pred}(\mathbb{X}\mathbf{R}+\mathbf{t}).
\end{equation*}
\vspace{-6mm}
\end{corollary}
% , Corollary~\ref{cor:equivariance} shows that the proposed network EqMotion is equivariant.

\section{Experiment}
\vspace{-1mm}
In this section, we validate our method on four different scenarios: particle dynamics, molecule dynamics, 3D human skeleton motion, and pedestrian trajectories. The detailed dataset description, implementation details, and additional experiment results are elaborated in the Appendix.

\textbf{Metric} We use: i) Average Displacement Error (ADE) and Final Displacement Error (FDE). ADE/FDE is the $\ell_2$ distance of the predicted whole motion/endpoint to the ground truth of the whole motion/endpoint; ii) Mean Per Joint Position Error (MPJPE). It records average $\ell_2$ distance between predicted joints and target ones at each future timestamp.

\begin{table}[t]
\setlength{\tabcolsep}{3pt}
\footnotesize
\centering
\renewcommand{\arraystretch}{1.1}
\caption{\small Interaction recognition accuracy and consistency (mean ± std in $\%$ in 5 independent runs) on the physical simulation.}
\vspace{-3mm}
\footnotesize
\setlength{\tabcolsep}{1mm}{
\begin{tabular}{c|cc|cc}
\hline
\hline
\rule{0pt}{10pt} % change row height

\multirow{2}{*}{\makecell[c]{\textbf{Model}}} & \multicolumn{2}{c|}{\makecell[c]{\textbf{Springs}}} & \multicolumn{2}{c}{\textbf{Charged}} \\
& Accuracy & Consistency & Accuracy  & Consistency\\
\hline
Corr.(path) \cite{kipf2018neural}& 58.1 {\scriptsize $\pm$ 0.0} & 99.8 {\scriptsize $\pm$ 0.1} &57.5 {\scriptsize $\pm$ 0.1} & 87.9 {\scriptsize $\pm$ 0.1}\\
Corr.(LSTM) \cite{kipf2018neural}& 53.5 {\scriptsize $\pm$ 0.5} & 92.4 {\scriptsize $\pm$ 2.1} &57.2 {\scriptsize $\pm$ 0.4} & 91.7 {\scriptsize $\pm$ 1.1}  \\
EGNN \cite{satorras2021n} & 61.0 {\scriptsize $\pm$ 1.3}& \textbf{100.0} {\scriptsize $\pm$ 0.0}&58.2 {\scriptsize $\pm$ 1.4}& \textbf{100.0} {\scriptsize $\pm$ 0.0}\\
NRI \cite{kipf2018neural}& 93.0 {\scriptsize $\pm$ 1.1} & 93.7 {\scriptsize $\pm$ 1.2} &70.0 {\scriptsize $\pm$ 0.6} & 88.5 {\scriptsize $\pm$ 1.3}\\
dNRI \cite{graber2020dynamic}& 93.3 {\scriptsize $\pm$ 2.0} & 89.6 {\scriptsize $\pm$ 2.0} &70.4 {\scriptsize $\pm$ 1.7} & 83.6 {\scriptsize $\pm$ 1.8} \\
\hline
\textbf{Ours} & \textbf{97.6} {\scriptsize $\pm$ 1.1} & \textbf{100.0} {\scriptsize $\pm$ 0.0} &\textbf{80.9} {\scriptsize $\pm$ 3.4} & \textbf{100.0} {\scriptsize $\pm$ 0.0}\\
Supervised & 98.7 {\scriptsize $\pm$ 0.2}& 100.0 {\scriptsize $\pm$ 0.0} & 97.4 $\pm$ 0.2 &100.0 {\scriptsize $\pm$ 0.0}\\
\hline
\hline
\end{tabular}}
\label{table:reasoning}
\vspace{-3.5mm}
\end{table}

\begin{figure}[t] 
\centering
\includegraphics[width=0.45\textwidth,height=0.22\textwidth]{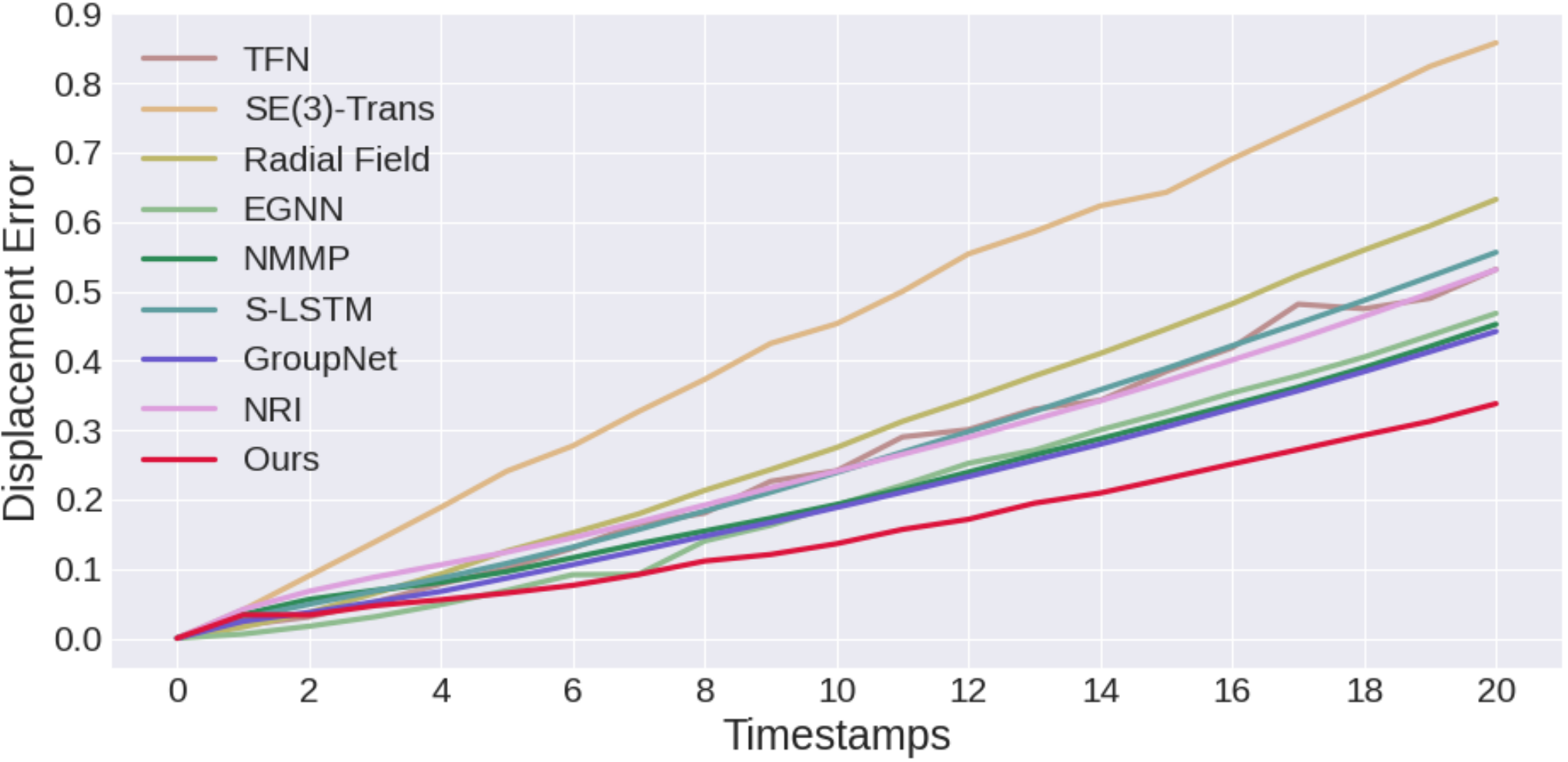}
\vspace{-3.5mm}
\caption{\small Comparison of displacement error across different timestamps on the physical simulation.}
\vspace{-4mm}
\label{fig:physical}
\end{figure}

\begin{table*}[t]
\vspace{-0.3cm}
\setlength{\tabcolsep}{2pt}
\caption{\small Comparisons of short-term skeleton motion prediction on 11 representative actions and average results across all actions on H3.6M.} 
\vspace{-0.3cm}
\renewcommand\arraystretch{0.9}
\resizebox{\textwidth}{!}{
\tiny
\begin{tabular}{c|cccc|cccc|cccc|cccc|cccc|cccc} \hline
Motion & \multicolumn{4}{c|}{Walking}                                   & \multicolumn{4}{c|}{Eating}                                    & \multicolumn{4}{c|}{Smoking}                                   & \multicolumn{4}{c}{Discussion} & \multicolumn{4}{c}{Directions}    & \multicolumn{4}{c}{Phoning}                             \\ \hline
millisecond     & 80           & 160          & 320          & 400 & 80           & 160          & 320          & 400    & 80          & 160         & 320         & 400         & 80          & 160         & 320         & 400         & 80          & 160          & 320          & 400          & 80           & 160          & 320          & 400          \\ \hline
Res-sup. (CVPR'17)         & 29.4          & 50.8          & 76.0          & 81.5          & 16.8          & 30.6          & 56.9          & 68.7          & 23.0          & 42.6          & 70.1          & 82.7          & 32.9          & 61.2          & 90.9          & 96.2    & 35.4          & 57.3          & 76.3          & 87.7              & 38.0          & 69.3          & 115.0         & 126.7       \\
Traj-GCN (ICCV'19)                & 12.3          & 23.0          & 39.8          & 46.1          & 8.4           & 16.9          & 33.2          & 40.7          &  7.9           &  16.2          & 31.9          & 38.9          & 12.5          & 27.4          & 58.5          & 71.7  & 9.0           & 19.9          & 43.4          &  53.7                  & 10.2          & 21.0          & 42.5          & 52.3        \\
DMGNN (CVPR'20)              & 17.3          & 30.7          & 54.6          & 65.2          & 11.0          & 21.4          & 36.2          & 43.9          & 9.0           & 17.6          & 32.1          & 40.3          & 17.3          & 34.8          & 61.0          & 69.8     & 13.1          & 24.6          & 64.7          & 81.9                & 12.5          & 25.8          & 48.1          & 58.3       \\
MSRGCN (ICCV'21)                &  12.2          &  22.7          &  38.6          &  45.2          &  8.4           & 17.1          &  33.0          &  40.4          & 8.0           & 16.3          &  31.3          &  38.2          &  12.0          &  26.8          &  57.1          &  69.7    &  8.6           &  19.7          &  43.3          & 53.8              &  10.1          &  20.7          &  41.5          &  51.3      \\
PGBIG (CVPR'22)                &   10.2          &   19.8          &   {34.5}          & {40.3}          & 7.0           & 15.1         & 30.6          &   {38.1}          &   {6.6}           &   {14.1}          &   {28.2}          &   {34.7}          &   {10.0}          &   {23.8}          &   {53.6}          &   {66.7}    &   {7.2}  &   {17.6} &   {40.9} &   {51.5}  &   {8.3}  &   {18.3} &   {38.7} &   {48.4}      \\ 
SPGSN (ECCV'22)                &   {10.1}          &   {19.4}          &   {34.8}          &   {41.5}          &   {7.1}           &   {14.9}          &   {30.5}          &   {37.9}          &   {6.7}           &   {13.8}          &   {28.0}          &   {34.6}          &   {10.4}          &   {23.8}          &   {53.6}          &   {67.1}    & 7.4          & 17.2          & 39.8          & 50.3                & 8.7 &18.3& 38.7& 48.5         \\
EqMotion (Ours) & \textbf{9.0}& \textbf{17.5}& \textbf{32.6}& \textbf{39.2}&\textbf{6.3}&\textbf{13.6}&\textbf{28.9}&\textbf{36.5}&\textbf{5.5}&\textbf{11.3}&\textbf{23.0}&\textbf{29.3}&\textbf{8.2}&\textbf{18.9}&\textbf{42.1}&\textbf{53.9}& \textbf{6.3}&\textbf{15.8}&\textbf{38.9}&\textbf{50.1}&\textbf{7.4}&\textbf{16.7}&\textbf{36.9}&\textbf{47.0}\\\hline  
% EqMotion (Ours) & \textbf{9.0}& \textbf{17.8}& 34.7& 41.5&\textbf{6.2}&\textbf{13.9}&\textbf{29.6}&\textbf{37.0}&\textbf{5.5}&\textbf{11.4}&\textbf{23.2}&\textbf{29.4}&\textbf{8.3}&\textbf{18.9}&\textbf{41.4}&\textbf{53.0}& \textbf{6.3}&\textbf{15.9}&\textbf{39.2}&\textbf{50.2}&\textbf{7.4}&\textbf{17.1}&\textbf{37.9}&\textbf{48.1}\\\hline         
Motion    & \multicolumn{4}{c|}{Posing}            & 
 \multicolumn{4}{c|}{Sitting} &
\multicolumn{4}{c|}{Sitting Down}                   & \multicolumn{4}{c|}{Waiting}    
% & \multicolumn{4}{c|}{Walking Dog}    
& \multicolumn{4}{c|}{Walking Together}                           & \multicolumn{4}{c}{Average}                                       \\ \hline
millisecond     & 80           & 160          & 320          & 400  & 80           & 160          & 320          & 400    & 80           & 160          & 320          & 400          & 80           & 160          & 320          & 400          & 80           & 160          & 320          & 400          & 80           & 160          & 320          & 400          \\ \hline
Res-sup. (CVPR'17)  & 36.1          & 69.1          & 130.5         & 157.1  & 42.6          & 81.4          & 134.7         & 151.8 &47.3&86.0&145.8&168.9        & 30.6          & 57.8          & 106.2         & 121.5                 & 26.8          & 50.1          & 80.2          & 92.2          & 34.7          & 62.0          & 101.1         & 115.5         \\
Traj-GCN (ICCV'19)  & 13.7          & 29.9          &  66.6          &  84.1   & 10.6          &  21.9          & 46.3          & 57.9  &16.1&31.1&61.5&75.5               & 11.4          & 24.0          & 50.1          & 61.5                   &  10.5          & 21.0          & 38.5          & 45.2          & 12.7          & 26.1          & 52.3          & 63.5          \\
DMGNN  (CVPR'20)  & 15.3          & 29.3          & 71.5          & 96.7 & 11.9          & 25.1          & 44.6          &   {50.2}  &15.0&32.9&77.1&93.0            & 12.2          & 24.2          & 59.6          & 77.5                  & 14.3          & 26.7          & 50.1          & 63.2          & 17.0          & 33.6          & 65.9          & 79.7          \\
MSRGCN (ICCV'21)    &  12.8          &  29.4          & 67.0          & 85.0  &  10.5          & 22.0          &  46.3          & 57.8    &16.1&31.6&62.5&76.8            &  10.7          &  23.1          &  48.3          & 59.2                    & 10.6          &  20.9          &  37.4         &  43.9          &  12.1          &  25.6          &  51.6          &  62.9         \\
PGBIG (CVPR'22)   &   {10.7} &   {25.7} &   {60.0} &   {76.6} &   {8.8}  &   {19.2} &   {42.4} &  53.8 & 13.9&27.9&57.4&71.5            &   {8.9}  &   {20.1} &   {43.6} &   {54.3}  &   {8.7}  &   {18.6} &   {34.4} &   {41.0} &   {10.3}  &   {22.7}  &   {47.4}  &   {58.5} \\ 
SPGSN (ECCV'22)    & 10.7& 25.3& 59.9& 76.5   & 9.3& 19.4 &{42.3}& {53.6} &14.2&27.7&56.8&\textbf{70.7}             & 9.2& 19.8& 43.1& 54.1                    & 8.9 &18.2& 33.8& 40.9        & 10.4 &22.3 &47.1& 58.3         \\ 
EqMotion (Ours)&\textbf{8.2}&\textbf{18.9}&\textbf{43.4}&\textbf{57.5}&\textbf{8.1}& \textbf{18.0} &\textbf{41.2}& \textbf{52.9}&\textbf{13.0}&\textbf{26.5}&\textbf{56.2}&\textbf{70.7}  &\textbf{7.6}& \textbf{17.4}& \textbf{39.9}& \textbf{51.1}&\textbf{7.8} &\textbf{16.1}& \textbf{30.6}& \textbf{37.1}& \textbf{9.1} &\textbf{20.1} &\textbf{43.7}& \textbf{55.0}\\\hline
\end{tabular}
}
\label{tab:human3.6_shortterm}
\vspace{-3.5mm}
\end{table*}

\begin{table*}[t]
\setlength{\tabcolsep}{3pt}
\caption{\small Comparisons of long-term skeleton motion prediction on 8 representative actions and average results across all actions on H3.6M.}
\vspace{-0.3cm}
\renewcommand\arraystretch{0.9}
\resizebox{\textwidth}{!}{
\begin{tabular}{c|cc|cc|cc|cc|cc|cc|cc|cc|cc} \hline
Motion & \multicolumn{2}{c|}{Walking}    & \multicolumn{2}{c|}{Eating}     & \multicolumn{2}{c|}{Smoking}     & \multicolumn{2}{c|}{Discussion}&  \multicolumn{2}{c|}{Greeting}   &  \multicolumn{2}{c|}{Phoning}         & \multicolumn{2}{c|}{Posing} & \multicolumn{2}{c|}{Walking Together}    & \multicolumn{2}{c}{Average}     \\ \hline
millisecond        & 560ms         & 1000ms         & 560ms         & 1000ms         & 560ms          & 1000ms         & 560ms          & 1000ms         & 560ms         & 1000ms         & 560ms          & 1000ms         & 560ms            & 1000ms           & 560ms          & 1000ms  & 560ms          & 1000ms       \\ \hline
Res-Sup. \cite{martinez2017human}         & 81.7          & 100.7          & 79.9          & 100.2          & 94.8           & 137.4          & 121.3          & 161.7  & 156.3          & 184.3                & 143.9            & 186.8            & 165.4          & 236.8 &173.6&202.3  & 129.2 & 165.0        \\
Traj-GCN \cite{mao2019learning}              & 54.1          & 59.8           & 53.4          & 77.8           & 50.7           & 72.6           & 91.6           & 121.5  &   {115.4}          & 148.8                  & 69.2             & 103.1            &   {114.5}          &   {173.0}  & 55.0&65.6   &81.6 & 114.3      \\
DMGNN \cite{li2020dynamic}            & 71.4          & 85.8           & 58.1          & 86.7           & 50.9           & 72.2           & 81.9           & 138.3      & 144.5          & 170.5            & 71.3             &   {108.4}             & 125.5          & 188.2   &70.5&86.9& 93.6 & 127.6        \\
MSRGCN \cite{dang2021msr}               &   {52.7}          &   {63.0}           &   {52.5}          &   {77.1}           &   {49.5}           &   {71.6}           &   {88.6}           &   {117.6}  & 116.3          &   {147.2}        &   {68.3}             & 104.4            & 116.3          & 174.3 &52.9&65.9   & 81.1 & 114.2      \\
PGBIG \cite{ma2022progressively}               &   {48.1} &   {56.4}  &   {51.1} &   {76.0}  &   {46.5}  &   {69.5}  &   {87.1}  &   {118.2} &   {110.2} &   {143.5}        &   {65.9}    &   {102.7}   &   {106.1} &   {164.8}&51.9&64.3 &76.9 & 110.3 \\ 
SPGSN \cite{li2022skeleton} &46.9 &53.6& 49.8& {73.4}& 46.7 &68.6& 89.7& 118.6& 111.0& {143.2}& 66.7& 102.5& 110.3& 165.4 &49.8&60.9& 77.4 & 109.6\\
% channel 96, seed 12
EqMotion (Ours) &\textbf{43.4} &\textbf{52.8}& \textbf{48.4}& \textbf{73.0}& \textbf{41.0} &\textbf{63.4}& \textbf{75.3}&\textbf{105.6}&\textbf{108.7}&\textbf{142.0}&\textbf{64.7}&\textbf{101.0} &\textbf{84.9}&\textbf{139.4} & \textbf{44.5} & \textbf{56.0}& \textbf{73.4} & \textbf{106.9}\\\hline
\end{tabular} 
}
\label{tab:Human3.6long-term}
\vspace{-6mm}
\end{table*}

\subsection{Scenario 1: Particle Dynamic Prediction}
\vspace{-1mm}
We use the particle $N$-body simulation \cite{kipf2018neural} in a 3D space similar to \cite{satorras2021n,fuchs2020se}. In the Springs simulation, particles are randomly connected by a spring. In the Charged simulation, particles are randomly charged or uncharged.  
% We predict the future motion of 20 timestamps given 20 timestamps past.  

\textbf{Validation on interaction reasoning} Since we can get the ground-truth interaction through simulation, we evaluate the ability of interaction reasoning as a category recognition task. 
To evaluate the robustness of the reasoning result with the Euclidean transformation, we introduce recognition "consistency", which is the ratio of the same recognition result under 20 random Euclidean transformations. 
The "Supervised" represents an upper bound that uses the ground-truth category to train the model. Table \ref{table:reasoning} reports the comparison of reasoning results on both springs and charged simulations,  for recognizing whether there is a spring connection or electrostatic force. We see that i) our method achieves significant improvement on the recognition accuracy and has more robust reasoning results due to the invariance of our reasoning module; 
and ii) our method achieves a close performance to the upper bound, reflecting that our method is capable of reasoning a robust and accurate interaction category.

\textbf{Validation on future prediction}
We conduct the experiment to evaluate the prediction performance on the charged settings. Figure \ref{fig:physical} compares the displacement error at all future timestamps of different methods. We see that our method (red line) achieves state-of-the-art prediction performance, reflecting our model's efficiency in future prediction.

\subsection{Scenario 2: Molecule Dynamic Prediction}
\vspace{-1mm}
We adopt the MD17 \cite{chmiela2017machine} dataset which contains the motions of different molecules generated via a
molecular dynamics simulation environment. The goal is to predict the motions of every atom of the molecule. We randomly pick four kinds of molecules: Aspirin, Benzene, Ethanol and Malonaldehyde and learn a prediction model for each molecule. 
% We predicted the future motion of 10 timestamps given the observation of 10 timestamps with a sampling rate of 20.
% We randomly pick 5k, 2k and 2k samples for training, validating and testing. Detail is presented in the Appendix. 

Table~\ref{tab:md17} presents the comparison of different motion prediction methods. We achieve state-of-the-art prediction performance on all four molecules. The ADE/FDE across four molecules is decreased by 34.2$\%$/30.1$\%$ on average.

\begin{table}[t]
  \centering
      \renewcommand\arraystretch{1}
    \setlength{\tabcolsep}{2.8pt}
  \caption{\small Prediction ADE/FDE ($\times 10^{-2}$) on the MD17 dataset.}
  \vspace{-3mm}
  % \resizebox{0.48\textwidth}{!}{
 \footnotesize
 % \scriptsize
    \begin{tabular}{ccccc}
    \toprule
          & \footnotesize{Aspirin} & \footnotesize{Benzene} & \footnotesize{Ethanol} & \footnotesize{Malonaldehyde} \\
    \midrule
    Radial Field \cite{kohler2019equivariant}& 17.98/26.20 & 7.73/12.47& 8.10/10.61 & 16.53/25.10    \\
    TFN  \cite{thomas2018tensor}&15.02/21.35&7.55/12.30&8.05/10.57&15.21/24.32  \\
    SE(3)-Trans \cite{fuchs2020se}&15.70/22.39&7.62/12.50&8.05/10.86&15.44/24.47    \\
    EGNN \cite{satorras2021n}&  14.61/20.65&7.50/12.16&8.01/10.22&15.21/24.00 \\
    LSTM  & 17.59/24.79 & 6.06/9.46 &7.73/9.88 &15.14/22.90\\
    S-LSTM \cite{alahi2016social}&13.12/18.14&3.06/3.52&7.23/9.85&11.93/18.43 \\
    NRI \cite{kipf2018neural}& 12.60/18.50 &1.89/2.58 & 6.69/8.78 &12.79/19.86 \\
    NMMP \cite{hu2020collaborative}&10.41/14.67&2.21/3.33&6.17/7.86&9.50/14.89 \\
    GroupNet \cite{xu2022groupnet}&10.62/14.00&2.02/2.95& 6.00/7.88&7.99/12.49 \\
    EqMotion(Ours) & \textbf{5.95}/\textbf{8.38}&\textbf{1.18}/\textbf{1.73}&\textbf{5.05}/\textbf{7.02}&\textbf{5.85}/\textbf{9.02}  \\
    \bottomrule
    \end{tabular}%
    % }
\vspace{-6mm}
  \label{tab:md17}%
\end{table}%

\subsection{Scenario 3: Human Skeleton Motion Prediction}
\vspace{-1mm}
We conduct experiments on the Human 3.6M (H3.6M) dataset \cite{ionescu2013human3} for 3D human skeleton motion prediction. H3.6M contains 7 subjects performing 15 actions. Following the standard paradigm \cite{martinez2017human,li2020dynamic}, we train the models on 6 subjects and tested on the specific clips of the 5th subject.

\textbf{Short-term motion prediction}
Short-term motion prediction aims to predict future poses within 400 milliseconds. We compare EqMotion with six state-of-the-art methods.
Table \ref{tab:human3.6_shortterm} presents the comparison of prediction MPJPEs. See the overall result in the Appendix. We see that i) EqMotion obtains superior performance at most timestamps across all the actions; and ii) compared to the baselines, EqMotion achieves much lower MPJPEs by 8.6$\%$ on average.

\textbf{Long-term motion prediction} 
Long-term motion prediction aims to predict the poses over 400 milliseconds. 
% We compare EqMotion with the same six state-of-the-art methods in short-term prediction. 
Table \ref{tab:Human3.6long-term} presents the prediction MPJPEs of various methods. We see that EqMotion achieves more effective prediction on most actions and has lower MPJPEs by 3.8 $\%$ on average. 

\begin{figure}[t] 
\centering
\includegraphics[width=0.47\textwidth]{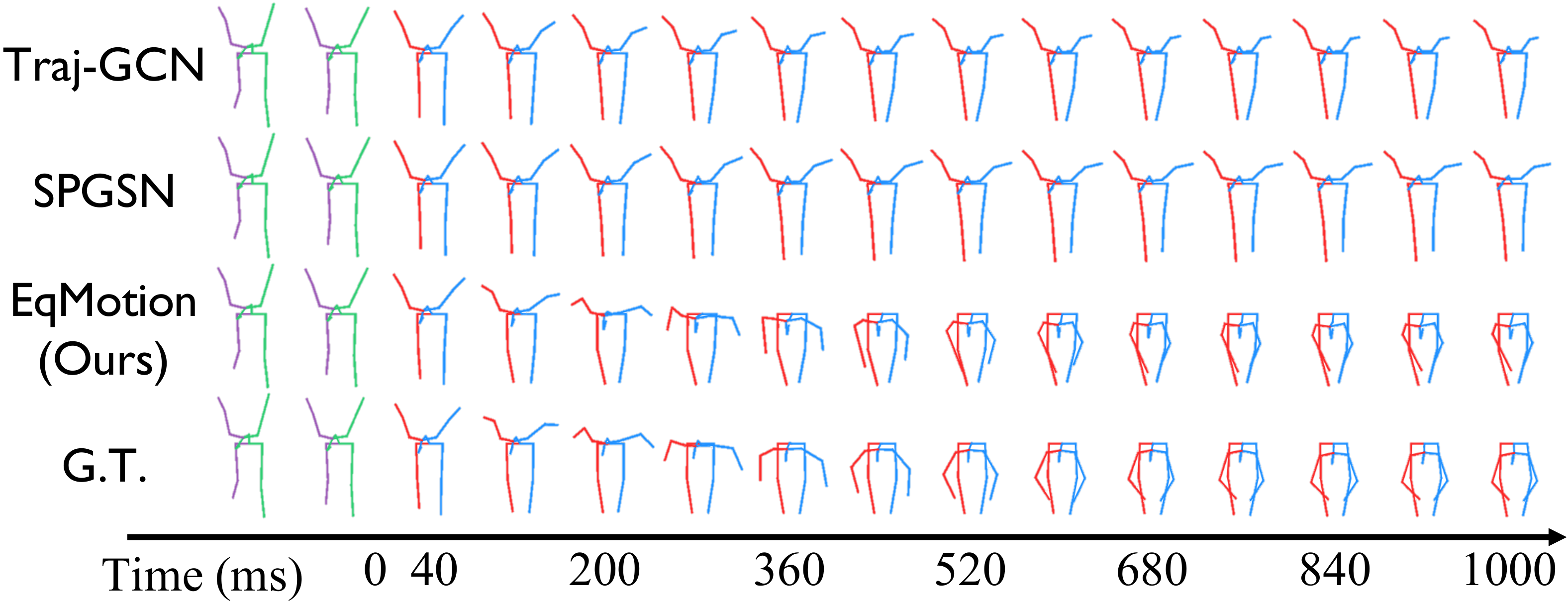}
\vspace{-3mm}
\caption{\small Qualitative comparison on the action ‘Sitting Down’ of H3.6M. EqMotion completes the action more accurately.
}
\vspace{-4mm}
\label{fig:vis}
\end{figure}

\textbf{Visualization results} Figure \ref{fig:vis} visualizes and compares the predictions on H3.6M. Baselines fail to predict future movements. EqMotion completes the action more precisely. 

The quantitative and visualization results reveal that EqMotion outperforms many previous methods that are task-specific, reflecting the effectiveness of EqMotion.

\subsection{Scenario 4: Pedestrian Trajectory Prediction}
\vspace{-1mm}
We conduct experiments on pedestrian trajectory prediction using the ETH-UCY dataset \cite{lerner2007crowds,pellegrini2009you}, which contains 5 subsets, ETH, HOTEL, UNIV, ZARA1, and ZARA2. 
% In the dataset, pedestrian trajectories are captured at 2.5Hz in multi-agent social scenarios. 
Following the standard setting \cite{alahi2016social,gupta2018social,yuan2021agentformer}, we use 3.2 seconds (8 timestamps) to predict the 4.8 seconds (12 timestamps). 
% We use the leave-one-out approach, training on 4 sets and testing on the remaining set.

Here we apply our EqMotion to two prediction modes: deterministic and multi-prediction. Deterministic means the model only outputs a single prediction for each input motion. Multi-prediction means the model has 20 predictions for each input motion. Under multi-prediction, ADE and FDE will be calculated by the best-performed predicted motion. To adapt to multi-prediction, we slightly modify EqMotion to repeat the last feature updating layer and the output layer 20 times in parallel to have a multi-head prediction, see the details in the Appendix. Table~\ref{table:eth} compares our method with sixteen baselines. We observe that i) EqMotion achieves state-of-the-art performance with the lowest average ADE and FDE, outperforming many baseline methods that are specifically designed for this task. EqMotion reduces the FDE by 10.4$\%$ and 7.9$\%$ under deterministic and multi-prediction settings; and ii) EqMotion achieves the best or the second best ADE/FDE result on most subsets, reflecting its effectiveness on the pedestrian trajectory prediction.

\begin{table}[t]
\begin{center}
\setlength{\tabcolsep}{2.3pt}
\caption{\small Prediction performance on the ETH-UCY dataset. The \textbf{bold}/\underline{underline} font denotes the best/second best result.}
\vspace{-3mm}
\scriptsize
% \begin{tabular}{c|c|c|c|c|c|c}
\begin{tabular}{c|ccccc|c}
% \hline %
\toprule
 & \multicolumn{6}{c}{Performance (ADE/FDE)} \\
% \hline
\midrule
Deterministic & ETH  & Hotel &Univ & Zara1 & Zara2  &Average\\
\midrule
% \hline
% Linear &1.33/2.94 &0.39/0.72 &0.62/1.21 & 0.77/1.48 & 0.82/1.59 & 0.79/1.59\\
% LSTM & 1.09/2.41 &0.86/1.91 & 0.61/1.31& 0.41/0.88 & 0.52/1.11 &0.70/1.52\\
S-LSTM \cite{alahi2016social}& 1.09/2.35 &0.79/1.76 & 0.67/1.40 & 0.47/1.00& 0.56/1.17 & 0.72/1.54\\
SGAN-ind \cite{gupta2018social} & 1.13/2.21 & 1.01/2.18 & 0.60/1.28 & 0.42/0.91 & 0.52/1.11 & 0.74/1.54\\
Traj++ \cite{salzmann2020trajectron++} &1.02/2.00&0.33/0.62 & 0.53/1.19 & 0.44/0.99 & 0.32/0.73 &0.53/1.11 \\
TransF \cite{giuliari2021transformer}&1.03/2.10&0.36/0.71 & 0.53/1.32 & 0.44/1.00 & 0.34/0.76 &0.54/1.17 \\
MemoNet \cite{xu2022remember} &1.00/2.08&0.35/0.67 & 0.55/1.19 & 0.46/1.00 & 0.37/0.82 &0.55/1.15 \\
EqMotion(Ours) & \textbf{0.96}/\textbf{1.92} & \textbf{0.30}/\textbf{0.58} & \textbf{0.50}/\textbf{1.10} &\textbf{0.39}/\textbf{0.86} & \textbf{0.30}/\textbf{0.68} &\textbf{0.49}/\textbf{1.03}\\
\midrule
% \hline
Multi-prediction & ETH  & Hotel &Univ & Zara1 & Zara2  &Average\\
\midrule
% \hline
SGAN \cite{gupta2018social}& 0.87/1.62 & 0.67/1.37 & 0.76/0.52 & 0.35/0.68 & 0.42/0.84 & 0.61/1.21\\
% STGAT~\cite{huang2019stgat} &0.65/1.12 & 0.35/0.66    & 0.34/0.69 & 0.29/0.60  & 0.52/1.10 & 0.43/0.83\\
NMMP \cite{hu2020collaborative} & 0.61/1.08 &
0.33/0.63& 0.52/1.11& 0.32/0.66& 0.43/0.85 &0.41/0.82 \\
% STAR \cite{yu2020spatio}& 0.36/0.65 &
% 0.17/0.36& 0.31/0.62 &0.29/0.52& 0.22/0.46 &0.26/0.53\\
Traj++ \cite{salzmann2020trajectron++}&0.61/1.02 & 0.19/0.28 & 0.30/0.54 & 0.24/0.42 & 0.18/0.31 & 0.30/0.51 \\
PECNet \cite{mangalam2020not}&0.54/0.87& 0.18/0.24 &0.35/0.60& 0.22/0.39& 0.17/0.30& 0.29/0.48 \\
Agentformer \cite{yuan2021agentformer}&0.45/0.75 &0.14/0.22 &0.25/\underline{0.45} &\underline{0.18}/\textbf{0.30} &\underline{0.14}/\underline{0.24}& \underline{0.23}/0.39 \\
GroupNet \cite{xu2022groupnet} & 0.46/0.73 & 0.15/0.25 & 0.26/0.49 & 0.21/0.39 & 0.17/0.33 & 0.25/0.44 \\
MID \cite{gu2022stochastic}& \textbf{0.39}/0.66 & \underline{0.13}/0.22 & \textbf{0.22}/\underline{0.45} &\textbf{0.17}/\textbf{0.30} & \textbf{0.13}/0.27 & \textbf{0.21}/\underline{0.38}\\
GP-Graph \cite{bae2022learning} &0.43/\underline{0.63} & 0.18/0.30 & {0.24}/0.42 & \textbf{0.17}/\underline{0.31} & 0.15/0.29 & 0.23/0.39 \\
EqMotion(Ours) &\underline{0.40}/\textbf{0.61} &\textbf{0.12}/\textbf{0.18}&\underline{0.23}/\textbf{0.43}&\underline{0.18}/{0.32}&\textbf{0.13}/\textbf{0.23}& \textbf{0.21}/\textbf{0.35} \\
% \hline
% \hline
\bottomrule
\end{tabular}
\label{table:eth}
\vspace{-7.5mm}
\end{center}
\end{table}

\begin{table}[t]
  \centering
    \renewcommand\arraystretch{0.9}
    \setlength{\tabcolsep}{3pt}
  \small
  \caption{\small Ablation study on key modules of EqMotion on H3.6M.}
    \vspace{-3mm}
  % \resizebox{0.48\textwidth}{!}{
  \small
    \begin{tabular}{ccc|cccccc}
    \toprule
        EGFL & IPFL & IRM  & 80ms & 160ms & 320ms & 400ms & Average \\
    \midrule
    & & & 12.9 & 31.9 & 68.2 & 82.4& 48.9 \\
      \checkmark& & &  10.1 &22.6 & 48.7& 60.7& 35.5 \\
       \checkmark& \checkmark& & 9.2 & 20.8& 45.4& 57.0& 33.1  \\
       \checkmark& \checkmark&\checkmark  & \textbf{9.1} &\textbf{20.1}& \textbf{43.7} &\textbf{55.0} & \textbf{32.0}\\
    \bottomrule
    \end{tabular}%
    % }
  \label{table:abltion_operation1}%
  \vspace{-3.5mm}
\end{table}

\begin{table}[t]
  \centering
      \renewcommand\arraystretch{0.9}
    \setlength{\tabcolsep}{5pt}
  \caption{\small Ablation study on operations in the equivariant geometric feature learning on H3.6M dataset.}
  \vspace{-3mm}
  % \resizebox{0.48\textwidth}{!}{
  \small
    \begin{tabular}{l|ccccc}
    \toprule
        Ablation  & 80ms & 160ms & 320ms & 400ms & Average \\
    \midrule
       w/o Inner att & 9.2 &20.5& 44.3&  55.7&  32.4  \\
       w/o Inter agg & 9.7 & 22.0 & 47.2 & 58.9 & 34.5 \\
        w/o Non-linear &  9.4 &21.3 &46.7 &58.6 & 34.0\\
       \textbf{EqMotion} & \textbf{9.1} &\textbf{20.1}& \textbf{43.7} &\textbf{55.0} & \textbf{32.0}\\
    
    \bottomrule
    \end{tabular}%
    % }
  \label{table:abltion_operation2}%
  \vspace{-6mm}
\end{table}

\vspace{-1mm}
\subsection{Ablation Studies}
\vspace{-1mm}
\textbf{Effect of network modules}
We explore the effect of three proposed key modules in EqMotion on the H3.6M dataset, including the equivariant geometric feature learning (EGFL), the invariant pattern feature learning (IPFL) and the invariant reasoning module (IRM). Table \ref{table:abltion_operation1} presents the experimental result. It is observed that i) the proposed three key modules all contribute to an accurate prediction; and ii) the equivariant geometric feature learning module is most important since learning a comprehensive equivariant geometric feature directly for prediction is the most important.

\textbf{Effect of equivariant operations}
We explore the effect of three proposed operations in the equivariant feature learning module in EqMotion, including the inner-agent attention (Inner att), inter-agent aggregation (Inter agg) and non-linear function (Non-linear). Table \ref{table:abltion_operation2} presents the results. We see that the proposed three key operations all contribute to promoting an accurate prediction.

\begin{figure}[t] 
\centering
\includegraphics[width=0.42\textwidth,height=0.24\textwidth]{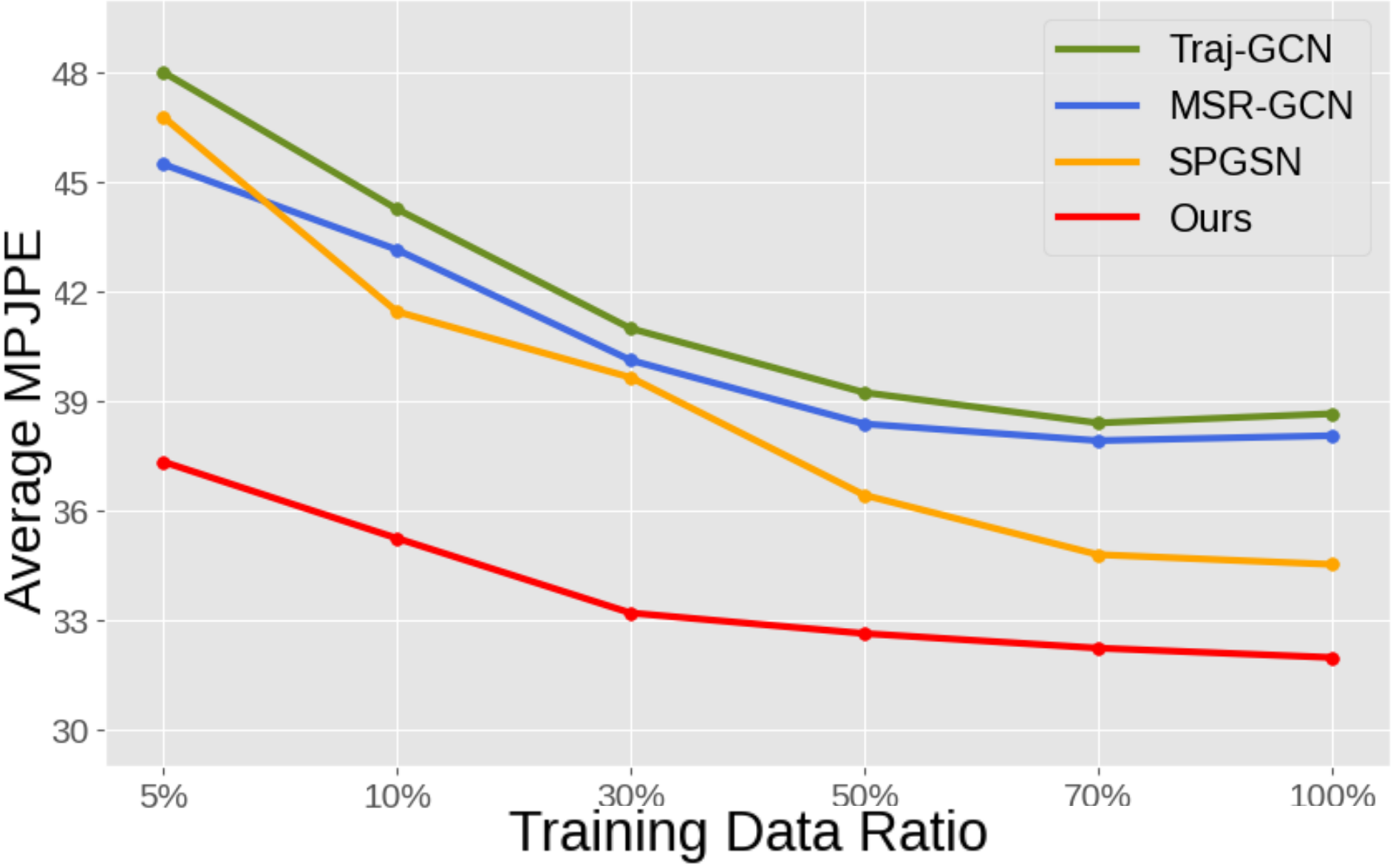}
\vspace{-3.5mm}
\caption{\small Comparison of model performance on different amounts of data in short-term prediction on H3.6M dataset.}
\vspace{-4.5mm}
\label{fig:data}
\end{figure}

\textbf{Different amounts of training data} Figure \ref{fig:data} presents the comparison of model performance under different amounts of training data on H3.6M. We see that i) our method achieves the best prediction performance under all training data ratios; and ii) our method even outperforms some full-data using baselines by only using 5$\%$ of training data since the equivariant design promotes the network generalization ability under Euclidean transformations.

\begin{figure}[t] 
\centering
\includegraphics[width=0.42\textwidth,height=0.24\textwidth]{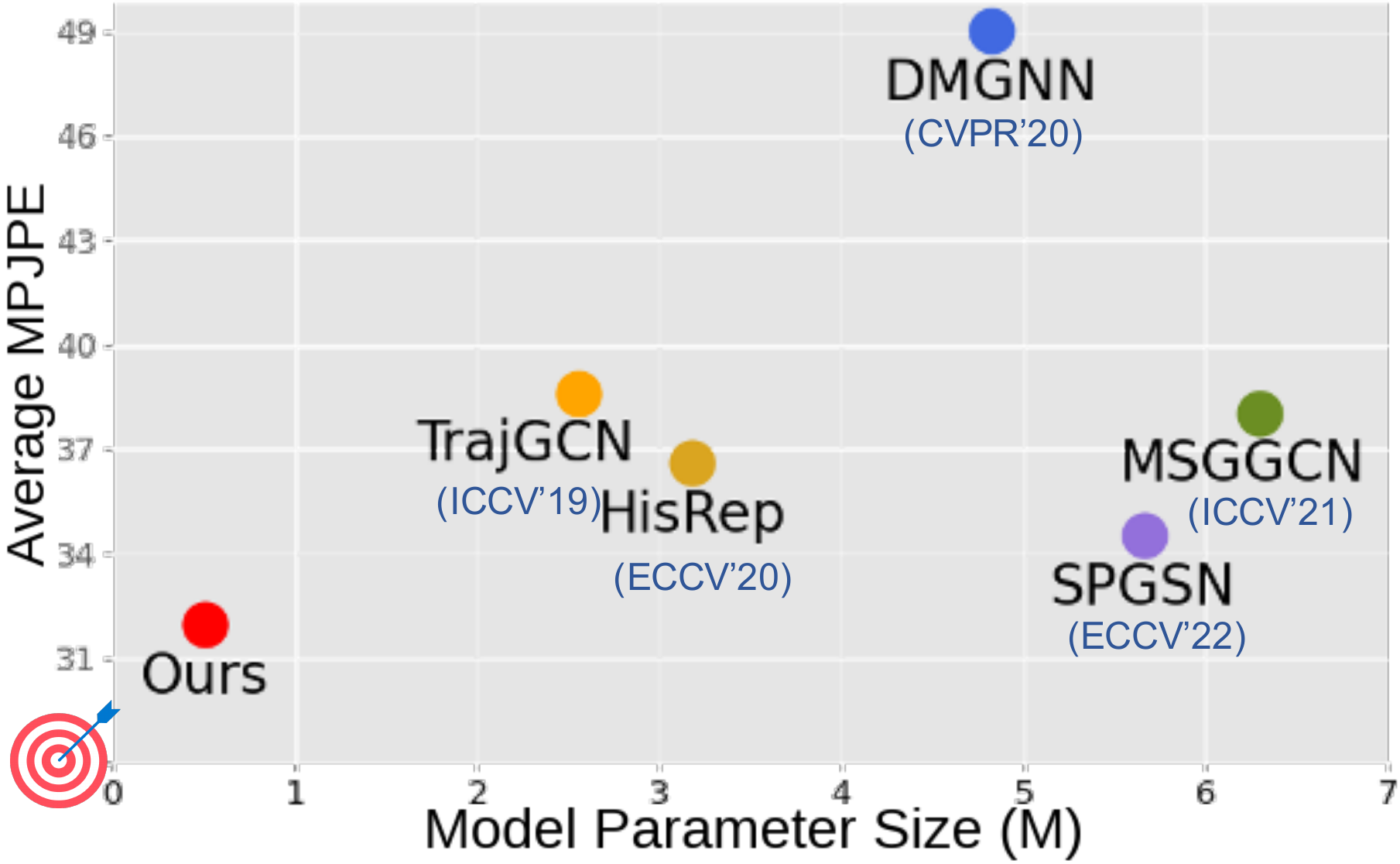}
\vspace{-3.5mm}
\caption{\small Comparison of model size and MPJPE in short-term prediction on H3.6M dataset. The target means the ideal model.}
\vspace{-5mm}
\label{fig:size}
\end{figure}

\textbf{Model size} Figure \ref{fig:size} compares EqMotion to existing methods in terms of the model size and prediction results in short-term prediction on H3.6M. We can observe that EqMotion has the smallest model size (less than 30\% of other models' sizes) with the lowest MPJPE thanks to the equivariant design that compacts the model free from generalizing over rotations and translations of the data.

\section{Conclusion}
\vspace{-1mm}
In this work, we present EqMotion, a motion prediction network that is theoretically equivariant under Euclidean transformations. EqMotion includes three novel designs: the equivariant geometric feature learning, the invariant pattern feature learning and the invariant reasoning module. We evaluate our method on four different scenarios and our method
achieves state-of-the-art prediction performance.
% \textbf{Limitation and future work}

\textbf{Acknowledgements.}
This project is partially supported by National Natural Science Foundation of China under Grant 62171276, the Science and Technology Commission of Shanghai Municipal under Grant 21511100900 and 22DZ2229005, and the National Research Foundation Singapore under its AI Singapore Programme (Award Number: AISG2-RP-2021-023). Robby T. Tan’s work is supported by MOE AcRF Tier, A-0009455-01-00.

%%%%%%%%% REFERENCES
{\small
\bibliographystyle{ieee_fullname}
\bibliography{egbib}
}

\newpage

% \and
\maketitle

% \tableofcontents

%%%%%%%%% BODY TEXT
\appendix
\section{Theoretical Proofs}
In this section, we prove Theorem 1 in our paper which shows EqMotion's equivariance property and the interaction reasoning module's invariance property. Note that, here we treat all the vectors to be row vectors since we multiply the rotation matrix by right.

% translation vector $\mathbf{t} \in \mathbb{R}^{n}$ and rotation (or reflection) matrix $\mathbf{R} \in {\rm SO}(n)$

1. For the initialization layer $\mathcal{F}_{\rm IL}(\cdot)$, the initial geometric feature is equivariant and the initial pattern feature is invariant:
\begin{equation*}
       \setlength{\abovedisplayskip}{2pt}
   \setlength{\belowdisplayskip}{2pt}
    \mathbb{G}^{(0)}\mathbf{R}+ \mathbf{t},\, \mathbf{H}^{(0)} = \mathcal{F}_{\rm IL}(\mathbb{X}\mathbf{R} + \mathbf{t}).
\end{equation*}
\textbf{Proof:} For the $i$th agent, we show its initial geometric feature is equivariant to the input motion under Euclidean transformation. When transforming the past motion, we have
\begin{equation}
\begin{aligned}
&\phi_{\rm init\_g}\big((\mathbf{X}_i\mathbf{R}+\mathbf{t})-\overline{\mathbb{X}}\mathbf{R}+\mathbf{t})\big)+\overline{\mathbb{X}} \mathbf{R}+\mathbf{t} \\
= \;&\mathbf{W}_{\rm init\_g}\big((\mathbf{X}_i\mathbf{R}-\overline{\mathbb{X}}\mathbf{R})\big)+\overline{\mathbb{X}}\mathbf{R}+\mathbf{t} \\
% &= \mathbf{W}_{\rm init\_g}((\mathbf{X}_i\mathbf{R}-\overline{\mathbb{X}}\mathbf{R}))+\overline{\mathbb{X}}\mathbf{R}+\mathbf{t}) \\
% &= \mathbf{W}_{\rm init\_g}\big((\mathbf{X}_i-\overline{\mathbb{X}})\big)\mathbf{R}+\overline{\mathbb{X}}\mathbf{R}+\mathbf{t} \\
= \;& (\mathbf{W}_{\rm init\_g}(\mathbf{X}_i-\overline{\mathbb{X}})+\overline{\mathbb{X}})\mathbf{R}+\mathbf{t} \\
=\;& \mathbb{G}^{(0)}\mathbf{R}+ \mathbf{t}
\label{eq:init1}
\end{aligned}
\end{equation}
Thus we show the initial geometric feature is equivariant to the input motion under Euclidean transformation. We also show its initial pattern feature is invariant to the input motion under Euclidean transformation. When transforming the past motion, we have,
\begin{equation}
\begin{aligned}
&\bigtriangleup (\mathbf{X}_i \mathbf{R}+\mathbf{t})=\bigtriangleup (\mathbf{X}_i)\mathbf{R} = \mathbf{V}_i \mathbf{R}, \\
&{||\mathbf{V}_i^t \mathbf{R}||_2^2}  = {\mathbf{V}_i^t}  \mathbf{R} \mathbf{R}^{\top} {\mathbf{V}_i^t}^{\top} = {\mathbf{V}_i^t}{\mathbf{V}_i^t}^{\top} = {||\mathbf{V}_i^t||_2^2} = \rho_i^t, \\
&\mathrm{angle}(\mathbf{V}_i^t \mathbf{R},\mathbf{V}_i^{t-1} \mathbf{R}) = \frac{\mathbf{V}_i^t \mathbf{R} (\mathbf{V}_i^{t-1} \mathbf{R})^\top }{||\mathbf{V}_i^t \mathbf{R}||_2 ||\mathbf{V}_i^{t-1} \mathbf{R}||_2} \\
& = \frac{\mathbf{V}_i^t \mathbf{R}\mathbf{R}^\top {\mathbf{V}_i^{t-1}}^\top}{||\mathbf{V}_i^t||_2 ||\mathbf{V}_i^{t-1}||_2}
=\frac{\mathbf{V}_i^t {\mathbf{V}_i^{t-1}}^\top}{||\mathbf{V}_i^t||_2 ||\mathbf{V}_i^{t-1}||_2} \\
&=\mathrm{angle}(\mathbf{V}_i^t,\mathbf{V}_i^{t-1}) = \theta_{i}^t, \\
% &\mathbf{V}_i=\bigtriangleup \mathbf{X}_i ,\rho_i^t = ||\mathbf{V}_i^t||_2, \theta_{i}^t = \mathrm{angle}(\mathbf{V}_i^t,\mathbf{V}_i^{t-1}), \\
%    & \mathbf{h}_i^{(0)}=\phi_{\mathrm {init\_h}}([\, \rho_i;\theta_i\,]),
&\phi_{\mathrm {init\_h}}([\, \rho_i;\theta_i\,])=\mathbf{h}_i^{(0)}.
\label{eq:init2}
\end{aligned}
\end{equation}
Thus we show the initial pattern feature is invariant to the input motion under Euclidean transformation.

% Combining Eq.\eqref{eq:init1} and Eq.\eqref{eq:init2}, we the initial geometric feature is equivariant and the initial pattern feature is invariant:

2. The reasoning module $\mathcal{F}_{\rm IRM}(\cdot)$ along with reasoned interaction categorical vectors $\{\mathbf{c}_{ij} \}$ is invariant:
\begin{equation*}
       \setlength{\abovedisplayskip}{2pt}
   \setlength{\belowdisplayskip}{2pt}
    \{\mathbf{c}_{ij} \} = \mathcal{F}_{\rm IRM}(\mathbb{G}^{(0)}\mathbf{R}+ \mathbf{t}, \mathbf{H}^{(0)}).
\end{equation*}

\noindent \textbf{Proof:} We first show the column-wise $\ell_2$-distance of geometric feature $||\mathbf{G}_i^{(0)}-\mathbf{G}_j^{(0)}||_{2,\mathrm {col}}$ is invariant since for the $c$th column ($c=1,\cdots,C$), we have $||\mathbf{g}_{i,c}^{(0)} \mathbf{R}+\mathbf{t}-(\mathbf{g}_{j,c}^{(0)} \mathbf{R}+\mathbf{t})||_2 = ||\mathbf{g}_{i,c}^{(0)} \mathbf{R}-\mathbf{g}_{j,c}^{(0)} \mathbf{R}||_2 = ||\mathbf{g}_{i,c}^{(0)}-\mathbf{g}_{j,c}^{(0)}||_2 $.
Since the initial pattern feature is invariant, thus we have the edge feature $\mathbf{m}_{ij}^{\prime}$, the aggregated edge feature $\mathbf{p}_i^{\prime}$ and the updated node feature $\mathbf{h}_i^\prime$ all to be invariant. Finally, we have the interaction categorical $\mathbf{c}_{ij}$ vector being invariant since $\mathbf{h}_i^\prime$ and $||\mathbf{G}_i^{(0)}-\mathbf{G}_j^{(0)}||_{2,\mathrm {col}}$ are invariant,
\begin{equation*}
       \setlength{\abovedisplayskip}{2pt}
   \setlength{\belowdisplayskip}{2pt}
  \mathrm{sm}\Big(\phi_{\rm rc}\big([\mathbf{h}_i^{\prime};\mathbf{h}_j^{\prime};||\mathbf{G}_i^{(0)}-\mathbf{G}_j^{(0)}||_{2,\mathrm {col}}]\big)/\tau\Big) =  \mathbf{c}_{ij}.
\end{equation*}

3. The $\ell$th geometric feature learning layer $\mathcal{F}_{\rm EGFL}^{(\ell)}(\cdot)$ is equivariant:
\begin{equation*}
       \setlength{\abovedisplayskip}{2pt}
   \setlength{\belowdisplayskip}{2pt}
\mathbb{G}^{(l+1)}\mathbf{R}+ \mathbf{t} = \mathcal{F}_{\rm EGFL}^{(\ell)}(\mathbb{G}^{(\ell)}\mathbf{R}+ \mathbf{t}, \mathbf{H}^{(\ell)},\{\mathbf{c}_{ij} \}).
\end{equation*}

\noindent \textbf{Proof:} We show the result by indicating the inner-agent attention, inter-agent aggregation and non-linear function all to be equivariant. We first show the inner-agent attention is equivariant. When transforming the input geometric feature, for every $i$th agent ($i=1,2,\cdots,M$),
\begin{equation}
\begin{aligned}
   % \mathbf{G}_i^{(\ell)} \leftarrow 
   &\phi^{(\ell)}_{\rm att}(\mathbf{h}_i^{(\ell)}) \cdot (\mathbf{G}_i^{(\ell)}\mathbf{R}+ \mathbf{t} -(\overline{\mathbb{G}}^{(\ell)}\mathbf{R}+ \mathbf{t}))+\overline{\mathbb{G}}^{(\ell)}\mathbf{R}+ \mathbf{t} \\
   = \;  &\phi^{(\ell)}_{\rm att}(\mathbf{h}_i^{(\ell)}) \cdot (\mathbf{G}_i^{(\ell)}- \overline{\mathbb{G}}^{(\ell)})\mathbf{R}+\overline{\mathbb{G}}^{(\ell)}\mathbf{R}+ \mathbf{t} \\
   = \; &(\phi^{(\ell)}_{\rm att}(\mathbf{h}_i^{(\ell)}) \cdot (\mathbf{G}_i^{(\ell)}- \overline{\mathbb{G}}^{(\ell)})+\overline{\mathbb{G}}^{(\ell)})\mathbf{R}+ \mathbf{t}\\
   \rightarrow \; &\mathbf{G}_i^{(\ell)}\mathbf{R}+ \mathbf{t}
   % \rightarrow \\ 
\end{aligned}
\end{equation}
Thus the inner-agent attention is equivariant. We then show the inter-agent aggregation is equivariant. When transforming the input geometric feature, we show the column-wise $\ell_2$-distance of geometric feature $||\mathbf{G}_i^{(\ell)}-\mathbf{G}_j^{(\ell)}||_{2,\mathrm {col}}$ is invariant since for the $c$th column ($c=1,\cdots,C$), we have $||\mathbf{g}_{i,c}^{(\ell)} \mathbf{R}+\mathbf{t}-(\mathbf{g}_{j,c}^{(\ell)} \mathbf{R}+\mathbf{t})||_2 = ||\mathbf{g}_{i,c}^{(\ell)} \mathbf{R}-\mathbf{g}_{j,c}^{(\ell)} \mathbf{R}||_2 = ||\mathbf{g}_{i,c}^{(\ell)}-\mathbf{g}_{j,c}^{(\ell)}||_2 $. Thus the learned aggregation weights $\mathbf{e}_{ij}^{(\ell)}$ is invariant. We have the inter-agent aggregation's equivariance,
\begin{equation}
\begin{aligned}
% &\mathbf{G}^{(\ell)}_i \leftarrow 
&\mathbf{G}^{(\ell)}_i \mathbf{R}+ \mathbf{t}+ \sum_{j \in \mathcal{N}_i} \mathbf{e}_{ij}^{(\ell)} \cdot (\mathbf{G}^{(\ell)}_i\mathbf{R}+ \mathbf{t}-(\mathbf{G}^{(\ell)}_j\mathbf{R}+ \mathbf{t})) \\
=\; & \mathbf{G}^{(\ell)}_i \mathbf{R}+ \mathbf{t}+ \sum_{j \in \mathcal{N}_i} \mathbf{e}_{ij}^{(\ell)} \cdot (\mathbf{G}^{(\ell)}_i-\mathbf{G}^{(\ell)}_j)\mathbf{R} \\
=\; & \big(\mathbf{G}^{(\ell)}_i + \sum_{j \in \mathcal{N}_i} \mathbf{e}_{ij}^{(\ell)} \cdot (\mathbf{G}^{(\ell)}_i-\mathbf{G}^{(\ell)}_j)\big)\mathbf{R}+ \mathbf{t} \\
\rightarrow \; & \mathbf{G}^{(\ell)}_i \mathbf{R}+ \mathbf{t}
\end{aligned}
\end{equation}
Thus the inner-agent attention is equivariant. We then show the non-linear function is equivariant. When transforming the input geometric feature, the inner product of the query coordinate and the key coordinate $\langle \mathbf{q}_{i,c}^{(\ell)}, \mathbf{k}_{i,c}^{(\ell)}\rangle = \mathbf{q}_{i,c}^{(\ell)}{\mathbf{k}_{i,c}^{(\ell)}}^\top $ is invariant for every channel $c=1,2,\cdots,C$ since
\begin{equation}
\begin{aligned}
&\mathbf{W}_{\rm Q}^{(\ell)}\big(\mathbf{G}_i^{(\ell)}\mathbf{R}+ \mathbf{t}-(\overline{\mathbb{G}}^{(\ell)}\mathbf{R}+ \mathbf{t} )\big) = \mathbf{Q}_i^{(\ell)}\mathbf{R}, \\
&\mathbf{W}_{\rm K}^{(\ell)}\big(\mathbf{G}_i^{(\ell)}\mathbf{R}+ \mathbf{t}-(\overline{\mathbb{G}}^{(\ell)}\mathbf{R}+ \mathbf{t} )\big) = \mathbf{K}_i^{(\ell)}\mathbf{R},\\
& \mathbf{Q}_i^{(\ell)}\mathbf{R}{(\mathbf{K}_i^{(\ell)}\mathbf{R})}^\top = \mathbf{Q}_i^{(\ell)}\mathbf{R}\mathbf{R}^\top{\mathbf{K}_i^{(\ell)}}^\top = \mathbf{Q}_i^{(\ell)}{\mathbf{K}_i^{(\ell)}}^\top.
\end{aligned}
\end{equation}
We also can have the equivariance of the two equations under different conditions,
\begin{equation}
\begin{aligned}
\mathbf{q}_{i,c}^{(\ell)} \mathbf{R}     +\overline{\mathbb{G}}^{(\ell)}\mathbf{R}+ \mathbf{t} = (\mathbf{q}_{i,c}^{(\ell)}      +\overline{\mathbb{G}}^{(\ell)})\mathbf{R}+ \mathbf{t} = \mathbf{g}_{i,c}^{(\ell+1)}\mathbf{R}+ \mathbf{t}
\end{aligned}
\end{equation}
and 
\begin{equation}
\begin{aligned}
&\mathbf{q}_{i,c}^{(\ell)}\mathbf{R} - \left\langle \mathbf{q}_{i,c}^{(\ell)}\mathbf{R}, \frac{\mathbf{k}_{i,c}^{(\ell)}\mathbf{R}}{||\mathbf{k}_{i,c}^{(\ell)}\mathbf{R}||_2}\right\rangle \frac{\mathbf{k}_{i,c}^{(\ell)}\mathbf{R}}{||\mathbf{k}_{i,c}^{(\ell)}\mathbf{R}||_2} +\overline{\mathbb{G}}^{(\ell)} \mathbf{R}+ \mathbf{t} \\
=\; & \mathbf{q}_{i,c}^{(\ell)}\mathbf{R} - \left\langle \mathbf{q}_{i,c}^{(\ell)}\mathbf{R}, \frac{\mathbf{k}_{i,c}^{(\ell)}\mathbf{R}}{||\mathbf{k}_{i,c}^{(\ell)}||_2}\right\rangle \frac{\mathbf{k}_{i,c}^{(\ell)}\mathbf{R}}{||\mathbf{k}_{i,c}^{(\ell)}||_2} +\overline{\mathbb{G}}^{(\ell)} \mathbf{R}+ \mathbf{t} \\
=\; & \mathbf{q}_{i,c}^{(\ell)}\mathbf{R} - \left\langle \mathbf{q}_{i,c}^{(\ell)}, \frac{\mathbf{k}_{i,c}^{(\ell)}}{||\mathbf{k}_{i,c}^{(\ell)}||_2}\right\rangle \frac{\mathbf{k}_{i,c}^{(\ell)}\mathbf{R}}{||\mathbf{k}_{i,c}^{(\ell)}||_2} +\overline{\mathbb{G}}^{(\ell)} \mathbf{R}+ \mathbf{t} \\
=\; & \left(\mathbf{q}_{i,c}^{(\ell)} - \left\langle \mathbf{q}_{i,c}^{(\ell)}, \frac{\mathbf{k}_{i,c}^{(\ell)}}{||\mathbf{k}_{i,c}^{(\ell)}||_2}\right\rangle \frac{\mathbf{k}_{i,c}^{(\ell)}}{||\mathbf{k}_{i,c}^{(\ell)}||_2} +\overline{\mathbb{G}}^{(\ell)}\right) \mathbf{R}+ \mathbf{t}\\
=\; & \mathbf{g}_{i,c}^{(\ell+1)}\mathbf{R}+ \mathbf{t}
\end{aligned}
\end{equation}
Since the criterion is invariant and two equations under two conditions are both equivariant, the non-linear function is equivariant.
Finally, combining the equivariance of inner-agent attention, inter-agent aggregation and nonlinear function, we show the equivariance of the geometric feature learning layer.

4. The $\ell$th pattern feature learning layer $\mathcal{F}_{\rm IPFL}^{(\ell)}(\cdot)$ is invariant:
\begin{equation*}
       \setlength{\abovedisplayskip}{2pt}
   \setlength{\belowdisplayskip}{2pt}
\mathbf{H}^{(l+1)} = \mathcal{F}_{\rm IPFL}^{(\ell)}(\mathbb{G}^{(\ell)}\mathbf{R}+ \mathbf{t}, \mathbf{H}^{(\ell)}).
\end{equation*}
\textbf{Proof:} Similar with the invariance of reasoning module, we first have the column-wise $\ell_2$-distance of geometric feature $||\mathbf{G}_i^{(\ell)}-\mathbf{G}_j^{(\ell)}||_{2,\mathrm {col}}$ is invariant.
Thus we have the variable in the message passing $\mathbf{m}_{ij}^{(\ell)}$, $\mathbf{p}_i^{(\ell)}$ all invariant. 
Finally the next layer's pattern feature $\mathbf{h}^{(l+1)}$ is invariant.
% $\mathbf{m}_{ij}^{(\ell)}$, the aggregated edge feature $\mathbf{p}_i^{(\ell)}$ and the updated node feature $\mathbf{h}_i^(\ell)$ all to be invariant.

5. The output layer $\mathcal{F}_{\rm EOL}(\cdot)$ is equivariant:
\begin{equation*}
       \setlength{\abovedisplayskip}{2pt}
   \setlength{\belowdisplayskip}{2pt}
\widehat{\mathbb{Y}}\mathbf{R}+ \mathbf{t} = \mathcal{F}_{\rm EOL}(\mathbb{G}^{(L)}\mathbf{R}+ \mathbf{t}).
\end{equation*}
\textbf{Proof:} When transforming the input geometric feature, 
\begin{equation}
    \begin{aligned}
&\mathcal{F}_{\rm EOL}(\mathbb{G}^{(L)}\mathbf{R}+ \mathbf{t}) \\
=\; & \big(\mathbf{W}_{\rm out}(\mathbf{G}_i^{(\ell)}\mathbf{R}+ \mathbf{t}-(\overline{\mathbb{G}}^{(\ell)}\mathbf{R}+ \mathbf{t})\big)+\overline{\mathbb{G}}^{(\ell)}\mathbf{R}+ \mathbf{t} \\
=\; & \big(\mathbf{W}_{\rm out}(\mathbf{G}_i^{(\ell)}-\overline{\mathbb{G}}^{(\ell)})+\overline{\mathbb{G}}^{(\ell)}\big)\mathbf{R}+ \mathbf{t} \\
=\; & \widehat{\mathbb{Y}}\mathbf{R}+ \mathbf{t}
    \end{aligned}
\end{equation}
Thus the output layer is equivariant.

\section{Optional Operations}
\subsection{DCT Processing}
To have a compact representation of input motion data, here we apply an optional discrete cosine transform (DCT) along the time axis to convert the input motion into the frequency domain. Mathematically, for the input motion $\mathbf{X}_i$ of agent $i$, we transform it by $\mathbf{X}_i \leftarrow \mathbf{W}_\mathrm{DCT}(\mathbf{X}_i-\overline{\mathbb{X}})$ where $\mathbf{W}_\mathrm{DCT} \in \mathbb{R}^{T_{\rm p}\times T_{\rm p}}$ is the DCT coefficients matrix. Correspondingly, we transform the predicted motion by an inverse DCT (iDCT) operation: $\widehat{\mathbf{Y}}_i \leftarrow \mathbf{W}_\mathrm{iDCT}\widehat{\mathbf{Y}}_i + \overline{\mathbb{X}}$ and $\mathbf{W}_\mathrm{iDCT} \in \mathbb{R}^{T_{\rm f}\times T_{\rm f}}$ is the iDCT coefficients matrix. The remove-and-add operation about the mean location $\overline{\mathbb{X}}$ is to ensure the translation equivariance. Since the DCT process is equivariant, adding this process will maintain whole network's equivariance.

\subsection{Adding Velocity Information}
We also introduce an optional operation to directly add the velocity information into the geometric feature by
\begin{equation}
    \begin{aligned}
        \mathbf{G}_i^{(\ell)} \leftarrow \phi_\rho(\rho_i) + \mathbf{G}_i^{(\ell)},
    \end{aligned}
\end{equation}
where $\rho_i$ is the velocity magnitude sequence and function $\phi_\rho(\cdot)$ is implemented by MLP. Since the velocity magnitude sequence is invariant, thus the operation is equivariant. This operation is placed before the nonlinear function. 

\section{Modification for Multi-prediction}
To make EqMotion perform multiple predictions in pedestrian trajectory prediction, we slightly modify the network by using multiple prediction heads in parallel. Each prediction head consists of a feature learning layer and an output layer. Assuming the $i$th output produced by the $i$th prediction head is $\widehat{\mathbb{Y}_i}$, we use a minimum $\ell_2$ prediction loss formulated by,
\begin{equation}
    \mathcal{L} = \mathop{min}\limits_{i}||\mathbb{Y}-\widehat{\mathbb{Y}_i}||^2_2.
\end{equation}
Through the loss, the optimal prediction will be optimized.

\section{Experiment Details}

\subsection{Dataset Description}

\subsubsection{Particle Dynamics}
We use the particle N-body simulation environment \cite{kipf2018neural} in a 3-dimensional space similar to \cite{satorras2021n,fuchs2020se}. The system contains 5 interacted particles. In the reasoning task, in the Springs simulation, particles will be randomly connected by a spring with a probability of 0.5. The particles connected by springs interact via forces given by Hooke’s law. In the Charged simulation, particles will be randomly charged or uncharged. The charged particles will repel or attract others via Coulomb forces. The probability of positive charged, uncharged and negative charged is 0.25, 0.5, and 0.25. We predicted the future motion of 20 timestamps given the historical observations of 20 timestamps. We use a downsampling rate of 100. We use 5k, 2k and 2k samples for training, validating and testing, respectively. In the prediction task, the setting is similar except the  probability of positive charged, uncharged and negative charged is 0.5, 0, and 0.5.

\subsubsection{Molecule Dynamics}
We adopt the MD17 \cite{chmiela2017machine} dataset which contains the motions of different molecules generated via a
molecular dynamics simulation environment. The goal is to predict the motions of every atom of the molecule. We randomly pick four kinds of molecules: Aspirin, Benzene, Ethanol and Malonaldehyde. We learn a prediction model for each molecule. 
We predicted the future motion of 10 timestamps given the observation of 10 timestamps. The raw data is a long sequence and we sample the trajectory with a sampling rate of 20 and a sampling gap of 400. We randomly pick 5k, 2k and 2k samples for training, validating and testing. 

\subsubsection{3D Human Skeleton Motion}
% We conduct experiments on the Human 3.6M (H3.6M) dataset \cite{ionescu2013human3} for 3D human skeleton motion prediction. H3.6M contains 7 subjects performing 15 actions. Following the standard paradigm \cite{martinez2017human,li2020dynamic}, we train the models on 6 subjects and tested on the specific clips of the 5th subject.
Human 3.6M (H3.6M) dataset \cite{ionescu2013human3} contains 7 subjects performing 15 classes of actions, and each subject has 22 body joints. All sequences are downsampled by two along time. Following previous paradigms \cite{martinez2017human,li2020dynamic}, the models are trained on the segmented clips in the 6 subjects and tested on the clips in the 5th subject.

\subsubsection{Pedestrian Trajectories}
ETH-UCY dataset \cite{lerner2007crowds,pellegrini2009you}, contains 5 subsets, ETH, HOTEL, UNIV, ZARA1, and ZARA2. 
In the dataset, pedestrian trajectories are captured at 2.5Hz in multi-agent social scenarios. 
Following the standard setting \cite{alahi2016social,gupta2018social,yuan2021agentformer}, we use 3.2 seconds (8 timestamps) to predict the 4.8 seconds (12 timestamps). 
We use the leave-one-out approach, training on 4 sets and testing on the remaining set.

% \begin{table}[t]
%   \centering
%     \setlength{\tabcolsep}{5pt}
%   \caption{\small Ablation study on operations in the equivariant geometric feature learning on short-term motion prediction on H3.6M dataset.}
%   \vspace{-3mm}
%   % \resizebox{0.48\textwidth}{!}{
%   \small
%     \begin{tabular}{l|ccccc}
%     \toprule
%         Ablation  & 80ms & 160ms & 320ms & 400ms & Average \\
%     \midrule
%        w/o Inner att & 9.2 &20.5& 44.3&  55.7&  32.4  \\
%        w/o Inter agg & 9.7 & 22.0 & 47.2 & 58.9 & 34.5 \\
%         w/o Non-linear &  9.4 &21.3 &46.7 &58.6 & 34.0\\
%        \textbf{EqMotion} & \textbf{9.1} &\textbf{20.1}& \textbf{43.7} &\textbf{55.0} & \textbf{32.0}\\
    
%     \bottomrule
%     \end{tabular}%
%     % }
%   \label{table:abltion_operation2}%
%   \vspace{-2mm}
% \end{table}

\subsection{Implementation Details}
In all the experiments, we set the number of feature learning layers $L$ to 4. We use the Adam optimizer to train the model on a single NVIDIA RTX-3090 GPU. All the MLPs have 2 layers with a ReLU activation function. 

\vspace{0.2cm}
\noindent\textbf{Particle Dynamics} We set the number of coordinates in the geometric feature $C$ as 64 and the dimension of the pattern feature $D$ as 64. The predefined category number $L$ is 2. We set the batch size to 50 and use a learning rate of 5e-4. The model is trained for 200 epochs. % with a decay rate of 0.95 for every 5 epochs.

\vspace{0.2cm}
\noindent\textbf{Molecule Dynamics} We set the number of coordinates in the geometric feature $C$ as 64 and the dimension of the pattern feature $D$ as 64. The predefined category number $L$ is 2. We set the batch size to 50 and use a learning rate of 5e-4. The model is trained for 300 epochs. % with a decay rate of 0.95 for every 5 epochs.

\vspace{0.2cm}
\noindent\textbf{Human Skeleton Motion} For short-term motion prediction, we set the number of coordinates in the geometric feature $C$ as 72 and the dimension of the pattern feature $D$ as 64. The predefined category number $L$ is 4. We set the batch size to 100 and use a learning rate of 5e-4. The model is trained for 80 epochs. % with a decay rate of 0.95 for every 5 epochs.
For long-term motion prediction, we set the number of coordinates in the geometric feature $C$ as 96 and the dimension of the pattern feature $D$ as 64. The predefined category number $L$ is 4. We set the batch size to 100 and use an initial learning rate of 5e-4 with a decay rate of 0.8 for every 2 epochs. The model is trained for 100 epochs.

\vspace{0.2cm}
\noindent\textbf{Pedestrian Trajectories} We set the number of coordinates in the geometric feature $C$ as 64 and the dimension of the pattern feature $D$ as 64. The predefined category number $L$ is 4. We set the batch size to 100 and use an initial learning rate of 8e-4/5e-4/1e-3/5e-4/1e-3 with a decay rate of 0.8/0.8/0.95/0.8/0.9 for every 2/2/2/2/2 epochs on eth/hotel/univ/zara1/zara2 subsets, respectively. The model is trained for 50 epochs.

\begin{table}[t]
  \centering
  \renewcommand\arraystretch{1.0}
    \setlength{\tabcolsep}{6pt}
  \footnotesize
  \caption{\small Effect of different numbers of learning layers on H3.6M.}
  \vspace{-3mm}
  % \resizebox{0.48\textwidth}{!}{
  \small
    \begin{tabular}{c|ccccc}
    \toprule
        Layers  & 80ms & 160ms & 320ms & 400ms & Average \\
    \midrule
       1 & 9.5 & 21.4 & 46.7 &58.3 & 34.0\\
       2 & 9.3 & 20.7 &45.4&56.5 & 33.0   \\
       3 & \textbf{9.1} & 20.3 & 44.3 & 55.7 & 32.4\\
       \textbf{4} & \textbf{9.1} &\textbf{20.1}& \textbf{43.7} &\textbf{55.0} & \textbf{32.0}\\
       5 & \textbf{9.1}& 20.2& 43.9 & 55.2 & 32.1\\
    
    \bottomrule
    \end{tabular}%
    % }
  \label{table:abltion_layers}%
  \vspace{-5mm}
\end{table}

\begin{table*}[t]
\vspace{-0.3cm}
\setlength{\tabcolsep}{3pt}
\caption{\small Comparisons of short-term prediction on Human3.6M. Results at 80ms, 160ms, 320ms, 400ms in the future are shown.}
%channel64 seed670
%python main_h36m_dyn.py --exp_name exp_1_dynegnn_vel_channel --model dynegnn_vel_channel --max_training_samples 5000 --lr 5e-4 --batch_size 100 --n_layers 4 --channels 64 --seed 810
\vspace{-0.3cm}
\renewcommand\arraystretch{0.9}
\resizebox{\textwidth}{!}{
\tiny
\begin{tabular}{c|cccc|cccc|cccc|cccc} \hline
Motion & \multicolumn{4}{c|}{Walking}                                   & \multicolumn{4}{c|}{Eating}                                    & \multicolumn{4}{c|}{Smoking}                                   & \multicolumn{4}{c}{Discussion}                                \\ \hline
millisecond        & 80          & 160         & 320         & 400         & 80          & 160         & 320         & 400         & 80          & 160          & 320          & 400          & 80           & 160          & 320          & 400          \\ \hline
Res-sup.          & 29.4          & 50.8          & 76.0          & 81.5          & 16.8          & 30.6          & 56.9          & 68.7          & 23.0          & 42.6          & 70.1          & 82.7          & 32.9          & 61.2          & 90.9          & 96.2          \\
Traj-GCN                & 12.3          & 23.0          & 39.8          & 46.1          & 8.4           & 16.9          & 33.2          & 40.7          &  7.9           &  16.2          & 31.9          & 38.9          & 12.5          & 27.4          & 58.5          & 71.7          \\
DMGNN              & 17.3          & 30.7          & 54.6          & 65.2          & 11.0          & 21.4          & 36.2          & 43.9          & 9.0           & 17.6          & 32.1          & 40.3          & 17.3          & 34.8          & 61.0          & 69.8          \\
MSRGCN                &  12.2          &  22.7          &  38.6          &  45.2          &  8.4           & 17.1          &  33.0          &  40.4          & 8.0           & 16.3          &  31.3          &  38.2          &  12.0          &  26.8          &  57.1          &  69.7          \\
PGBIG                &   10.2          &   19.8          &   {34.5}          & {40.3}          & 7.0           & 15.1         & 30.6          &   {38.1}          &   {6.6}           &   {14.1}          &   {28.2}          &   {34.7}          &   {10.0}          &   {23.8}          &   {53.6}          &   {66.7}          \\ 
SPGSN                &   {10.1}          &   {19.4}          &   {34.8}          &   {41.5}          &   {7.1}           &   {14.9}          &   {30.5}          &   {37.9}          &   {6.7}           &   {13.8}          &   {28.0}          &   {34.6}          &   {10.4}          &   {23.8}          &   {53.6}          &   {67.1}          \\
EqMotion(Ours) & \textbf{9.0}& \textbf{17.5}& \textbf{32.6}& \textbf{39.2}&\textbf{6.3}&\textbf{13.6}&\textbf{28.9}&\textbf{36.5}&\textbf{5.5}&\textbf{11.3}&\textbf{23.0}&\textbf{29.3}&\textbf{8.2}&\textbf{18.8}&\textbf{42.1}&\textbf{53.9}\\\hline
Motion           & \multicolumn{4}{c|}{Directions}                                & \multicolumn{4}{c|}{Greeting}                                  & \multicolumn{4}{c|}{Phoning}                                   & \multicolumn{4}{c}{Posing}                                    \\ \hline
millisecond        & 80           & 160          & 320          & 400          & 80           & 160          & 320          & 400          & 80           & 160          & 320          & 400          & 80           & 160          & 320          & 400          \\ \hline
Res-sup.          & 35.4          & 57.3          & 76.3          & 87.7          & 34.5          & 63.4          & 124.6         & 142.5         & 38.0          & 69.3          & 115.0         & 126.7         & 36.1          & 69.1          & 130.5         & 157.1         \\
Traj-GCN                & 9.0           & 19.9          & 43.4          &  53.7          & 18.7          & 38.7          & 77.7          & 93.4          & 10.2          & 21.0          & 42.5          & 52.3          & 13.7          & 29.9          &  66.6          &  84.1          \\
DMGNN              & 13.1          & 24.6          & 64.7          & 81.9          & 23.3          & 50.3          & 107.3         & 132.1         & 12.5          & 25.8          & 48.1          & 58.3          & 15.3          & 29.3          & 71.5          & 96.7          \\
MSRGCN                &  8.6           &  19.7          &  43.3          & 53.8          &  16.5          &  37.0          &  77.3          &  93.4          &  10.1          &  20.7          &  41.5          &  51.3          &  12.8          &  29.4          & 67.0          & 85.0          \\
PGBIG                &   {7.2}  &   {17.6} &   {40.9} &   {51.5} &   {15.2} &   {34.1} &   {71.6} &   {87.1} &   {8.3}  &   {18.3} &   {38.7} &   {48.4} &   {10.7} &   {25.7} &   {60.0} &   {76.6} \\ 
SPGSN               & 7.4          & 17.2          & 39.8          & 50.3          & 14.6         & 32.6          & {70.6}         & {86.4}         & 8.7 &18.3& 38.7& 48.5        & 10.7& 25.3& 59.9& 76.5               \\ 
EqMotion(Ours) & \textbf{6.3}&\textbf{15.8}&\textbf{38.9}&\textbf{50.1}&\textbf{12.7}&\textbf{30.1}&\textbf{68.3}&\textbf{85.2}&\textbf{7.4}&\textbf{16.7}&\textbf{36.9}&\textbf{47.0}&\textbf{8.2}&\textbf{18.9}&\textbf{43.4}&\textbf{57.5} \\ \hline
Motion           & \multicolumn{4}{c|}{Purchases}                                 & \multicolumn{4}{c|}{Sitting}                                   & \multicolumn{4}{c|}{Sittingdown}                               & \multicolumn{4}{c}{Takingphoto}                               \\ \hline
millisecond        & 80           & 160          & 320          & 400          & 80           & 160          & 320          & 400          & 80           & 160          & 320          & 400          & 80           & 160          & 320          & 400          \\ \hline
Res-sup.          & 36.3          & 60.3          & 86.5          & 95.9          & 42.6          & 81.4          & 134.7         & 151.8         & 47.3          & 86.0          & 145.8         & 168.9         & 26.1          & 47.6          & 81.4          & 94.7          \\
Traj-GCN                & 15.6          & 32.8          & 65.7          &  79.3          & 10.6          &  21.9          & 46.3          & 57.9          & 16.1          &  31.1          &  61.5          &  75.5          & 9.9           &  20.9          & 45.0          & 56.6          \\
DMGNN              & 21.4          & 38.7          & 75.7          & 92.7          & 11.9          & 25.1          & 44.6          &   {50.2}          & 15.0          & 32.9          & 77.1          & 93.0          & 13.6          & 29.0          & 46.0          & 58.8          \\
MSRGCN                &  14.8          &  32.4          & 66.1          & 79.6          &  10.5          & 22.0          &  46.3          & 57.8          &  16.1          & 31.6          & 62.5          & 76.8          &  9.9           & 21.0          &  44.6          &  56.3         \\
PGBIG                &   {12.5} &   {28.7} &   \textbf{60.1} &   \textbf{73.3} &   {8.8}  &   {19.2} &   {42.4} &  53.8 &   {13.9} &   {27.9} &   {57.4} &   {71.5} &   {8.4}  &   {18.9} &   {42.0} &   {53.3} \\ 
SPGSN                & 12.8 &28.6 &61.0& 74.4         & 9.3& 19.4 &{42.3}& {53.6}          & 14.2 &27.7& 56.8& \textbf{70.7}       & 8.8& 18.9& \textbf{41.5}& \textbf{52.7}              \\ 
EqMotion(Ours) &\textbf{11.2} &\textbf{26.8} &{60.5} &75.2&\textbf{8.1}& \textbf{18.0} &\textbf{41.2}& \textbf{52.9}&\textbf{13.0}& \textbf{26.5}& \textbf{56.2}& \textbf{70.7}&\textbf{7.9}& \textbf{17.7}& 40.9& 52.8 \\\hline
Motion           & \multicolumn{4}{c|}{Waiting}                                   & \multicolumn{4}{c|}{Walking Dog}                                & \multicolumn{4}{c|}{Walking Together}                           & \multicolumn{4}{c}{Average}                                       \\ \hline
millisecond        & 80           & 160          & 320          & 400          & 80           & 160          & 320          & 400          & 80           & 160          & 320          & 400          & 80           & 160          & 320          & 400          \\ \hline
Res-sup.          & 30.6          & 57.8          & 106.2         & 121.5         & 64.2          & 102.1         & 141.1         & 164.4         & 26.8          & 50.1          & 80.2          & 92.2          & 34.7          & 62.0          & 101.1         & 115.5         \\
Traj-GCN                & 11.4          & 24.0          & 50.1          & 61.5          & 23.4          & 46.2          & 83.5          & 96.0          &  10.5          & 21.0          & 38.5          & 45.2          & 12.7          & 26.1          & 52.3          & 63.5          \\
DMGNN              & 12.2          & 24.2          & 59.6          & 77.5          & 47.1          & 93.3          & 160.1         & 171.2         & 14.3          & 26.7          & 50.1          & 63.2          & 17.0          & 33.6          & 65.9          & 79.7          \\
MSRGCN                &  10.7          &  23.1          &  48.3          & 59.2          &  20.7          &  42.9          &  80.4          &  93.3          & 10.6          &  20.9          &  37.4         &  43.9          &  12.1          &  25.6          &  51.6          &  62.9         \\
% Ours                &   {8.9}  &   {20.1} &   {43.6} &   {54.3} &   {18.8} &   {39.3} &   {73.7} &   {86.4} &   {8.7}  &   {18.6} &   {34.4} &   {41.0} &   {10.3} (\textcolor{blue}{-1.8}) &   {22.7} (\textcolor{blue}{-2.9}) &   {47.4} (\textcolor{blue}{-4.2}) &   {58.5} (\textcolor{blue}{-4.4}) \\ \hline

PGBIG                &   {8.9}  &   {20.1} &   {43.6} &   {54.3} &   {18.8} &   {39.3} &   {73.7} &   {86.4} &   {8.7}  &   {18.6} &   {34.4} &   {41.0} &   {10.3}  &   {22.7}  &   {47.4}  &   {58.5} \\ 
SPGSN                & 9.2& 19.8& 43.1& 54.1        &  17.8& 37.2& \textbf{71.7}& \textbf{84.9}               & 8.9 &18.2& 33.8& 40.9        & 10.4 &22.3 &47.1& 58.3         \\ 
EqMotion(Ours) &\textbf{7.6}& \textbf{17.4}& \textbf{39.9}& \textbf{51.1}& \textbf{16.6}&\textbf{36.4} &{72.5}& {86.2}&\textbf{7.8} &\textbf{16.1}& \textbf{30.6}& \textbf{37.1}& \textbf{9.1} &\textbf{20.1} &\textbf{43.7}& \textbf{55.0}\\\hline
\end{tabular}
}
\label{tab:supp_human3.6_shortterm}
\vspace{-3mm}
\end{table*}

\begin{table*}[ht]
\caption{\small Comparisons of long-term prediction on Human3.6M. Results at 560ms and 1000ms in the future are shown.}
\vspace{-0.3cm}
\renewcommand\arraystretch{0.9}
\resizebox{\textwidth}{!}{
\begin{tabular}{c|cc|cc|cc|cc|cc|cc|cc|cc} \hline
Motion & \multicolumn{2}{c|}{Walking}    & \multicolumn{2}{c|}{Eating}     & \multicolumn{2}{c|}{Smoking}     & \multicolumn{2}{c|}{Discussion}  & \multicolumn{2}{c|}{Directions} & \multicolumn{2}{c|}{Greeting}    & \multicolumn{2}{c|}{Phoning}         & \multicolumn{2}{c}{Posing}      \\ \hline
millisecond        & 560ms         & 1000ms         & 560ms         & 1000ms         & 560ms          & 1000ms         & 560ms          & 1000ms         & 560ms         & 1000ms         & 560ms          & 1000ms         & 560ms            & 1000ms           & 560ms          & 1000ms         \\ \hline
Res-Sup.          & 81.7          & 100.7          & 79.9          & 100.2          & 94.8           & 137.4          & 121.3          & 161.7          & 110.1         & 152.5          & 156.3          & 184.3          & 143.9            & 186.8            & 165.7          & 236.8          \\
Traj-GCN                & 54.1          & 59.8           & 53.4          & 77.8           & 50.7           & 72.6           & 91.6           & 121.5          &   {71.0}          & 101.8          &   {115.4}          & 148.8          & 69.2             & 103.1            &   {114.5}          &   {173.0}          \\
DMGNN              & 71.4          & 85.8           & 58.1          & 86.7           & 50.9           & 72.2           & 81.9           & 138.3          & 102.1         & 135.8          & 144.5          & 170.5          & 71.3             &   {108.4}             & 125.5          & 188.2          \\
MSRGCN                &   {52.7}          &   {63.0}           &   {52.5}          &   {77.1}           &   {49.5}           &   {71.6}           &   {88.6}           &   {117.6} & 71.2          &   {100.6}          & 116.3          &   {147.2}          &   {68.3}             & 104.4            & 116.3          & 174.3          \\
PGBIG                &   {48.1} &   {56.4}  &   {51.1} &   {76.0}  &   {46.5}  &   {69.5}  &   {87.1}  &   {118.2}          &  \textbf{69.3} &   \textbf{100.4} &   {110.2} &   {143.5} &   {65.9}    &   {102.7}   &   {106.1} &   {164.8} \\ 
SPGSN &46.9 &53.6& 49.8& {73.4}& 46.7 &68.6& 89.7& 118.6& 70.1& 100.5& 111.0& {143.2}& 66.7& 102.5& 110.3& 165.4 \\
% channel 96, seed 12
EqMotion(Ours)  &\textbf{43.4} &\textbf{52.8}& \textbf{48.4}& \textbf{73.0}& \textbf{41.0} &\textbf{63.4}& \textbf{75.3}&\textbf{105.6}&70.4&101.3&\textbf{108.7}&\textbf{142.0}&\textbf{64.7}&\textbf{101.0} &\textbf{84.9}&\textbf{139.4}\\\hline
Motion          & \multicolumn{2}{c|}{Purchases}  & \multicolumn{2}{c|}{Sitting}    & \multicolumn{2}{c|}{Sitting Down} & \multicolumn{2}{c|}{Taking Photo} & \multicolumn{2}{c|}{Waiting}    & \multicolumn{2}{c|}{Walking Dog}  & \multicolumn{2}{c|}{Walking Together} & \multicolumn{2}{c}{Average}     \\ \hline
millisecond        & 560ms         & 1000ms         & 560ms         & 1000ms         & 560ms          & 1000ms         & 560ms          & 1000ms         & 560ms         & 1000ms         & 560ms          & 1000ms         & 560ms            & 1000ms           & 560ms          & 1000ms         \\ \hline
Res-Sup.          & 119.4         & 176.9          & 166.2         & 185.2          & 197.1          & 223.6          & 107.0          & 162.4          & 126.7         & 153.2          & 173.6          & 202.3          & 94.5            & 110.5            & 129.2        & 165.0          \\
Traj-GCN                & 102.0         & 143.5          & 78.3          & 119.7          &   {100.0}          &   {150.2}          &   {77.4}           &   {119.8}          & 79.4          & 108.1          & 111.9          & 148.9          & 55.0             &   {65.6}             & 81.6           & 114.3          \\
DMGNN              & 104.9         & 146.1          &   {75.5}          &   {115.4}          & 118.0          & 174.1          & 78.4           & 123.7          & 85.5         & 113.7          & 183.2          & 210.2          & 70.5             & 86.9            & 93.6          & 127.6          \\
MSRGCN                &   {101.6}         &   {139.2}          & 78.2          & 120.0          & 102.8          & 155.5          & 77.9           & 121.9          &   {76.3}          &   {106.3}          &   {111.9}          &   {148.2}          &   {52.9}             & 65.9             &   {81.1}           &   {114.2}          \\
% Ours                &   {95.3} &   {133.3} &   {74.4} &   {116.1} &   {96.7}  &   {147.8} &   {74.3}  &   {118.6} &   {72.2} &   {103.4} &   {104.7} &   {139.8} &   {51.9}    &   {64.3}    &   {76.9} (\textcolor{blue}{-4.2})  &   {110.3} (\textcolor{blue}{-3.9}) \\ \hline

PGBIG                &  {95.3} &   \textbf{133.3} &   \textbf{74.4} &   \textbf{116.1} &    \textbf{96.7}  &   \textbf{147.8} &   \textbf{74.3}  &   {118.6} &   \textbf{72.2} &   \textbf{103.4} &   {104.7} &   {139.8} &   {51.9}    &   {64.3}    &   {76.9}  &   {110.3} \\ 
SPGSN& 96.5& 133.9& 75.0& 116.2& 98.9& 149.9& 75.6& \textbf{118.2}& 73.5& 103.6& \textbf{102.4}& \textbf{138.0}& 49.8& 60.9& 77.4& 109.6\\
EqMotion(Ours) &\textbf{93.5}&134.5&74.7&116.6&{98.1}&{149.9}&76.7&122.0&71.4&104.6&104.8&141.2&\textbf{44.5}&\textbf{56.0}&\textbf{73.4}&\textbf{106.9}\\\hline

\end{tabular} 
}
\label{tab:supp_Human3.6long-term}
\end{table*}

\section{Further Experiment Results}

\noindent\textbf{Different numbers of layers}
Table \ref{table:abltion_layers} shows the effect of different numbers of feature learning layers $L$ on the H3.6M dataset. We find that i) initially increasing $L$ leads to better performance as a more comprehensive geometric feature and pattern feature will be learned; and ii) when the number of layers is sufficient, the performance tends to be stable.

\section{Limitation and Future Work}
This work focuses on a generally applicable motion prediction method. In the future, we plan to expand the method by adding specific designs for different tasks to further improve the model performance. We also expect the method can use more types of data to assist prediction, such as images and videos that contain map information.

\end{document}